\documentclass{article}
\usepackage{amssymb, amsmath, amsthm, bm}
\usepackage[noend]{algpseudocode}
\usepackage{algorithmicx,algorithm}
\usepackage{graphicx}
\usepackage{subfigure}
\usepackage{caption}
\usepackage{bbding}
\usepackage{booktabs}
\usepackage{tikz,xcolor}
\usepackage{float}
\usepackage[section]{placeins}
\usepackage[affil-it]{authblk}
\newtheorem{rmrk}{Remark}

\title{PI-VEGAN: Physics Informed Variational Embedding Generative Adversarial Networks for Stochastic Differential Equations}

\author[1]{Ruisong Gao\thanks{grs1108@163.com}}
\author[1]{Yufeng Wang\thanks{ytuyufengwang@163.com}}
\author[1]{Min Yang\thanks{yang@ytu.edu.cn}}
\author[1]{Chuanjun Chen\thanks{cjchen@ytu.edu.cn}}
\affil[1]{School of Mathematics and Information Sciences, Yantai University, Yantai, China}
\date{\today}

\begin{document}
\maketitle

\begin{abstract}
We present a new category of physics-informed neural networks called physics informed variational embedding generative adversarial network (PI-VEGAN),
that effectively tackles the forward, inverse, and mixed problems of stochastic differential equations.
In these scenarios, the governing equations are known, but only a limited number of sensor measurements of the system parameters are available.
We integrate the governing physical laws into PI-VEGAN with automatic differentiation,
while introducing a variational encoder for approximating the latent variables of the actual distribution of the measurements.
These latent variables are integrated into the generator to facilitate accurate learning of the characteristics of the stochastic partial equations.
Our model consists of three components, namely the encoder, generator, and discriminator,
each of which is updated alternatively employing the stochastic gradient descent algorithm.
We evaluate the effectiveness of PI-VEGAN in addressing forward, inverse, and mixed problems that require the concurrent calculation of system parameters and solutions.
Numerical results demonstrate that the proposed method achieves satisfactory stability and accuracy in comparison with the previous physics-informed generative adversarial network (PI-WGAN).
\end{abstract}

\section{Introduction}
Stochastic differential equations (SDEs)  arise in many fields, including finance, physics, and engineering, and typically involve random fluctuations in the underlying system.
The computational  methods used to solve these equations must be able to handle both the spatial and temporal variability of the problem,
as well as the stochastic nature of the solution.

Over the past four decades, there has been a significant increase in the development of numerical methods for stochastic differential equations.
Monte Carlo methods, which involve simulating the underlying stochastic process using random sampling techniques, are commonly used to solve SDEs.
Other numerical methods for SDEs include spectral methods, which use Fourier or wavelet transforms to solve the equation in the frequency domain,
and stochastic Galerkin methods, which involve projecting the SDE onto a finite-dimensional space and solving the resulting system of equations using standard numerical techniques.
For instance, Kloeden and Platen \cite{Kloeden1989} examined various strong and weak approximation methods based on the stochastic Taylor formula for Ito stochastic differential equations.
Stability and implementation issues have been discussed in \cite{Burrage2000}, with variable step size implementation shown to be superior to fixed step size for small stochasticity.
Tocino \cite{Tocino2002} developed a class of explicit Runge-Kutta schemes of second order in the weak sense for systems of stochastic differential equations with multiplicative noise.
Furthermore, two Runge-Kutta schemes of third order were obtained for scalar equations with constant diffusion coefficients,
where the first method generated independent identically distributed approximations of the solution by sampling the coefficients of the equation
and used a standard Galerkin finite element variational formulation.
Babuska et. al. \cite{Babuska2004} proposed a finite dimensional approximation of the stochastic coefficients,
turning the original stochastic problem into a deterministic parametric elliptic problem.
A Galerkin finite element method of either the $h$- or $p$-version was then used to approximate the corresponding deterministic solution,
which led to approximations of the desired statistics.
However, traditional numerical methods may not be efficient for high dimensional stochastic partial equations and can suffer from the "curse of dimensionality".

In recent years, the use of deep learning to solve fundamental partial differential equations (PDEs) has gained considerable attention  \cite{Feng2022,Sheng2022,Wilson2022},
thanks to the high expressiveness of neural networks and the rapid growth of computing hardware.
Among them,
physics-informed neural networks (PINNs) \cite{Chen2021,Gao2023,Jagtap2022,Jagtap2020,Karumuri2021,Liu2021,Raissi2019,Xu2023,Yang2019,Yang2022} are a particularly interesting approach.
PINNs incorporate physical knowledge as soft constraints in the empirical loss function
and employ machine learning methodologies like automatic differentiation and stochastic optimization to train the model.
In \cite{Nabian2019}, the random PDE is approximated by a feed-forward deep residual network, with either strong or weak enforcement of initial and boundary constraints.
A reinforcement learning method was  presented in \cite{E2017} to solve backward stochastic differential equations,
where the gradient of the solution plays the role of a policy function and the loss function is given by the error between the prescribed terminal condition and the solution.
Chen et. al. \cite{Chen2019} employed a Karhunen–Lo\`{e}ve expansion for the stochastic diffusivity and arbitrary polynomial chaos for the solution,
and then designed multiple neural networks to solve  stochastic advection–diffusion–reaction systems.
A machine learning method, lifting the requirement for a deterministic forward solver, was presented by \cite{Karumuri2020} for high-dimensional uncertainty propagation of elliptic SDEs.
In many practical problems, it is impossible to access to exact analytical representations of the parameters of equations
and can only obtain limited information about the parameters through sparse sensors.
To address these scenarios, 
Guo et al. \cite{Guo2022} used the normalized field flow method to tackle data-driven stochastic differential equations.
In particular, it is noteworthy that  Liu et al. \cite{Liu2020} introduced PI-WGAN, 
which utilizes the powerful generation capability of generative adversarial networks (GANs) \cite{Goodfellow2014} to solve forward, inverse, and mixed problems in a unified manner.
However, the inputs to the generator in PI-WGAN were sampled from a prior distribution that is independent of the distribution of the collected measurements,
thereby leading to unstable performance sometimes.

In this paper, we present a novel approach to improving the accuracy and training stability of PI-WGAN by leveraging the implicit distribution information contained in collected measurements.
Our proposed method involves developing a physics-informed neural network that incorporates the governing physical laws into its architecture using automatic differentiation \cite{Baydin2018},
and introducing a variational encoder to produce the latent variables of the real distribution of the measurements.
Our model is comprised of three parts: encoder, generator, and discriminator, and the corresponding training procedure are outlined as follows.
First, the encoder receives real snapshots and outputs the corresponding latent variables.
These latent variables, along with the spatial coordinates, are then fed into the generator to produce synthetic snapshots.
The discriminator takes in both the real and generated snapshots and distinguishes between them.
We utilize the well-known variational inference technique \cite{Kingma2013} to optimize the embedded encoder,
and thus name our approach Physics-Informed Variational Embedding Generative Adversarial Network (PI-VEGAN).
PI-VEGAN is a flexible and efficient approach capable of solving a wide range of forward, inverse, and mixed problems.
Since the inputs to the generator are learned from the training samples rather than being manually determined,
they can provide valuable guidance to the GAN's learning process,
potentially leading to better convergence and overall performance than PI-WGAN in solving stochastic differential equations.
It is worth noting that the variational technique has also been employed in a recent study by Zhong et al. \cite{Zhong2023}. 
However, their method is based on the VAE framework, which does not incorporate a discriminator and has distinct optimization objectives compared to our approach.

The rest of this article is organized as follows.
In Section 2, we describe the problem to be solved  and then briefly review the relevant models.
In Section 3, we introduce PI-VEGAN in detail.
We show how to fit stochastic processes by the proposed method, and further how to solve stochastic differential equations.
Numerical experiments are provided in Section 4.
The limitation of the approach and the prospects for future work are discussed in Section 5.

\section{Background}
\subsection{Problem Setup}
Consider the following stochastic differential equation:
\begin{align}
	\label{eq1}
	\begin{split}
		\mathcal{N}_x[u(x; \omega),k(x; \omega)] & = f(x; \omega),    	
		\quad x \in \mathcal{D},\quad\omega \in \Omega,
		\\[5pt]
		\mathcal{B}_x[u(x; \omega)] & = b(x; \omega),
		\quad x \in \Gamma,
	\end{split}
\end{align}
where $\mathcal{N}_x$ denotes a general differential operator,
$\mathcal{D}$ is a physical domain in $\mathbb{R}^d$,
$x$  is the spatial coordinate,
$\Omega$  is a probability space,
$\omega$ is a random event,
and $\mathcal{B}_x$ denotes the operator acting on the domain boundary $\Gamma$.
Since the coefficient $ k(x; \omega)$ ,  the forcing term $ f(x; \omega)$  and the boundary condition $ b(x;\omega) $ are random processes,
then the latent solution $ u(x; \omega)$ is also a random process depend on $ k(x; \omega)$ and $ f(x; \omega)$.
We can collect a number of measurements through the scattered sensors for the random processes in (\ref{eq1}).

Without loss of generality,
assume that the coordinates of the sensors for $ u(x; \omega)$, $ k(x; \omega)$, $ f(x; \omega)$ and  $b(x; \omega)$ are $\{x_i^u\}_{i=1}^{n_u}$, $\{x_i^k\}_{i=1}^{n_k}$,  $\{x_i^f\}_{i=1}^{n_f}$ and $\{x_i^b\}_{i=1}^{n_b}$, respectively,
where $n_u$, $n_k$, $n_f$ and $n_b$ denote the number of the sensors.
Depending of the available measurements,
three types of problems are derived, namely, forward,  inverse, and mixed problems.

\textbf{Forward problem} is to approximate the solution $u(x; \omega) $ given the measurements of $ k(x; \omega)$, $ f(x; \omega)$ and  $b(x; \omega)$.

\textbf{Inverse problem} is to estimate the coefficient $ k(x; \omega)$ given the measurements of $ u(x; \omega)$, $ f(x; \omega)$ and  $b(x; \omega)$.

\textbf{Mixed problem}  is to compute the solution $ u(x; \omega)$ and the coefficient $ k(x; \omega)$ simultaneously with only partial knowledge of $ u(x; \omega)$ and $ k(x; \omega)$ available.
As the number of sensors on $ u(x; \omega)$ increases from zero while the number of sensors on $ k(x; \omega)$ decreases,
the estimation gradually changes from a forward problem to a mixed problem, and finally to an inverse problem.

Given a random event $\omega \in \Omega$,
we can use the scattered sensors to collect the snapshots of the  stochastic processes.	
Suppose we have a group of $N$ snapshots denoted by
\begin{align}
	\label{RealSnap}
	\{H(\omega^{(j)})\}_{j=1}^N = \{(K(\omega^{(j)}), U(\omega^{(j)}), F(\omega^{(j)}), B(\omega^{(j)}))\}_{j=1}^N,
\end{align}
where
\begin{align*}
	K(\omega^{(j)}) &= (k(x_i^k; \omega^{(j)}))_{i=1}^{n_k}, \quad
	U(\omega^{(j)}) = (u(x_i^u; \omega^{(j)}))_{i=1}^{n_u},
	\\[5pt]
	F(\omega^{(j)}) &= (f(x_i^f; \omega^{(j)}))_{i=1}^{n_f},\quad
	B(\omega^{(j)}) = (b(x_i^b; \omega^{(j)}))_{i=1}^{n_b}.
\end{align*}
The corresponding terms in \eqref{RealSnap} are omitted if we put no sensors for that process.
In the forward problem $n_u =0 $, and in the inverse problem $ n_k =1 $.

\subsection{Generative Adversarial Network}
Generative adversarial networks (GANs) are a type of deep learning models that employ adversarial training strategy for both the generator and discriminator.
During the training process, the generator $\mathcal{G}$ endeavors to fabricate data to deceive the discriminator,
whereas the discriminator $\mathcal{D}$ attempts to distinguish between the real data and the ones generated by the generator.
The two networks engage in a dynamic game process, and the ultimate optimization objective is expressed as follows \cite{Goodfellow2014}:
\begin{align}
	\min_{\mathcal{G}}\mathop{\max}_{\mathcal{D}}\mathbb{E}_{y }[\log \mathcal{D}(y)]
	+ \mathbb{E}_{\tilde{z}}[\log(1 - \mathcal{D}(\mathcal{G}(\tilde{z})))],
\end{align}
where $ \tilde{z} $ is the noise sampled from a given prior distribution (e.g. Gaussian distribution),
and $ y $ denotes the real sample.

Due to the highly unstable training process of vanilla GAN,
WGAN-GP \cite{Gulrajani2017} was introduced as a solution, aimed at mitigating this issue.
The approach involves the implementation of weight clipping and gradient penalization techniques.
The optimization objective for the WGAN-GP approach is outlined below:
\begin{align}
	\label{WGAN}
	\min_{\mathcal{G}}\mathop{\max}_{\mathcal{D}}\mathbb{E}_{y}[\mathcal{D}(y)] - \mathbb{E}_{\tilde{z}}[\mathcal{D}(\mathcal{G}(\tilde{z}))]
	+ \lambda (\|\nabla_{\hat{y}}\mathcal{D}(\hat{y})\|_2 - 1)^2.
\end{align}
Here $\hat{y}=\epsilon y + (1 - \epsilon)\tilde{y}$,
$\epsilon$ follows the uniform distribution $ U(0, 1)$,
 $\tilde{y}=\mathcal{G}(\tilde{z})$,
and $\lambda$ is a regularization hyperparameter.

Daw et al. \cite{Daw2021} used GAN to perform the uncertainty quantization problem.
Based on WGAN-GP,
Liu et. al. \cite{Liu2020} built a physics informed neural network  named PI-WGAN for solving stochastic partial differential equations.
However, the inputs to the generator are sampled from a predetermined distribution that  is independent of the distribution of the collected measurements,
which makes the performance of PI-WGAN is not very stable.

\subsection{Variational Inference}

To enhance the stability of PI-WGAN, we aim to propose a modified architecture that includes an auxiliary encoder network.
The encoder, denoted by $\mathcal{E}$, extracts the distribution information of the samples by mapping the sample $y$ to a latent variable space with a posterior distribution $p(z|y)$.
However, estimating the posterior distribution $p(z|y)$ directly is not advisable, given the heavy computational cost involved.

The  idea of variational inference \cite{Kingma2013} is to define a surrogate distribution $q(z)$ with some variational parameters,
and then compute the optimal setting of these parameters in order to make $q(z)$ as close as possible to the posterior distribution $p(z|y)$.
The closeness of the two distributions can be measured by Kullback-Leibler (KL) divergence,
which is defined as
\begin{align}
	\textrm{KL}(q||p(z|y))=  \int_{z}q(z) \log \frac{q(z)}{p(z|y)}dz.
\end{align}
The smaller the KL diverengece, the closer the two distributions are.

It is impossible to minimize the KL divergence exactly because $ p(z|y) $ is unknown.
However, note that
\begin{align*}				
	\textrm{KL}(q||p(z|y)) &= \int_{z}q(z) \log q(z)dz - \int_z q(z) \log p(z|y) dz
	\\[5pt]
	&=  \int_{z}q(z) \log q(z)dz - \int_z q(z) \log p(z,y) dz +\log p(y)
	\\[5pt]
	&=  \int_{z}q(z) \log q(z)dz - \int_z q(z) \log p(y|z)p(z) dz +\log p(y).
\end{align*}
Therefore, minimize the KL divergence is the same as minimizing
\begin{align*}				
	\int_{z}q(z) \log q(z)dz -\int_z q(z) \log p(y|z)p(z) dz
	=\textrm{KL}(q||p(z))-\mathbb{E}_{q}[\log p(y|z)].
\end{align*}

We can typically specify that the function $p(z)$ conforms to a $d$-dimensional Gaussian distribution $N(0, I_d)$.
Additionally, we require that the surrogate function $q(z)$ follows a normal distribution $N(\mu, \sigma I_d)$,
where the parameters $\mu$ and $\sigma$ are outputs of the encoder $\mathcal{E}$.
With knowledge of $\mu$ and $\sigma$, we can derive the latent variable as $z = \mu + \sigma\odot\xi$, where $\xi \sim N(0, I_d)$.

We argue that feed the learned latent variable to the generator in equation \eqref{WGAN} can lead to more stable and accurate  solution.

\section{Our Approach}

In this section, we first use a stochastic process to illustrate the basic framework of PI-VEGAN.
Then we provide the complete procedure of PI-VEGAN in solving stochastic differential equations.

\subsection{Approximation of stochastic processes}

Suppose that we have collected a group of snapshots for a  stochastic process  $ f(x; \omega) $:
\begin{align}
	\{F(\omega^{(j)})\}_{j=1}^N = \{(f(x_i; \omega^{(j)}))_{i=1}^{n_f}\}_{j=1}^N,
\end{align}
where $ N $ is the number of  snapshots,  $n_f$ is the number of scattered sensors and $\{x_i\}_{i=1}^{n_f}$ are locations of the sensors.

We use a generator network $\tilde{f}_\theta(x; z) $ parameterized by $ \theta $ to model the stochastic process $ f(x; \omega) $.
Let $\mathcal{E}_{\phi}(\cdot)$ and $\mathcal{D}_{\rho}(\cdot)$ be the encoder and discriminator,
parameterized by $\phi$ and $ \rho $, respectively.

First, the encoder $\mathcal{E}_{\phi}(\cdot)$, which consists of $n_f$ input neurons, takes $ F(\omega^{(j)})$ as input and outputs the mean $\mu_j \in \mathbb{R}^d$ and variance $\sigma_j \in \mathbb{R}^d$ of the latent variable.
Define the latent variable $ z_j \in \mathbb{R}^d $ as
\begin{align}
	\label{z_sample_1}
	z_j = \mu_j + \sigma_j \odot \xi_j, \quad  \xi_j \sim N(0, I_d),
\end{align}
where $ N(0,I_d) $ denotes the $d$-dimensional Gaussian distribution.

Then the generator $\tilde{f}_\theta$ takes the concatenation of  the spatial coordinate $x_i$ and the latent variable $z_j$ as the input,
and generates a ``fake" sample $ \tilde{f}_\theta(x_i;z_j) $.
So we have $N$ ``fake" snapshots denoted by
\begin{align}
	\label{F}
	\{\tilde{F}(\omega^{(j)})\}_{j=1}^N = \{( \tilde{f}_\theta (x_i;z_j) )_{i=1}^{n_f}\}_{j=1}^N.
\end{align}

Finally, the discriminator $\mathcal{D}_{\rho}(\cdot)$ takes the ``fake" or real snapshot as input and tries to distinguish between them.
\begin{rmrk}
 Note that from \eqref{F} that in the training procedure, the inputs $\{z_j\} $ to generator are learned from the real snapshots distribution,
 rather than being sampled from a predetermined distribution such as the standard normal distribution.
 Such approach enables the inputs to provide valuable guidance to the GAN's learning process, 
 potentially leading to improved convergence and overall performance in approximating stochastic processes.
\end{rmrk}

Given positive loss weights $\alpha$, $\eta$, $\gamma$, and $ \lambda $,
the optimization objective  for modeling the stochastic process $ f(x; \omega) $ consists of the following three components:
\begin{align}
\label{loss_E_1}			
	\min_{\phi}\textrm{KL}(q(z|F(\omega^{(j)}))||p(z)) + \frac{1}{N} \sum_{j=1}^{N} [ \tilde{F}(\omega^{(j)}) - F(\omega^{(j)}) ]^2.
	\quad p(z) \sim N(0,I_d),
\end{align}

\begin{align}
\label{loss_G_1}
	\min_{\theta}\frac{1}{N}\sum_{j=1}^{N}\bigg(- \alpha \mathcal{D}_{\rho}(\tilde{F}(\omega^{(j)}))
	+ \eta [\tilde{F}(\omega^{(j)})-F(\omega^{(j)})]^2 \bigg),
\end{align}

\begin{align}
\label{loss_D_1}
	\min_{\rho}\frac{1}{N}\sum_{j=1}^{N}\bigg(\gamma [\mathcal{D}_{\rho}(\tilde{F}(\omega^{(j)}))
	- \mathcal{D}_{\rho}(F(\omega^{(j)}))]
	+ \lambda \mathcal{L}_{pen}(\tilde{F}(\omega^{(j)}))\bigg),
\end{align}
where $ \mathcal{L}_{pen}(\tilde{F}(\omega^{(j)}) $  is a regularization term satisfying \cite{Gulrajani2017}
\begin{align*}
	\mathcal{L}_{pen} = \frac{1}{N}\sum_{j=1}^{N} \big(\|\nabla_{\hat{F}(\omega^{(j)})}\mathcal{D}_{\rho}(\hat{F}(\omega^{(j)}))\| - 1 \big)^2
\end{align*}
with $ 	\hat{F}(\omega^{(j)})= (1 - \epsilon_{j})\tilde{F}(\omega^{(j)})+\epsilon_{j} F(\omega^{(j)})$, and $\epsilon_j \sim U(0, 1)$.

In \eqref{loss_E_1} and \eqref{loss_G_1},
a squared loss  $(\tilde{F}(\omega^{(j)})-F(\omega^{(j)}))^2$ is introduced to force the output $\tilde{F}(\omega^{(j)})$  to reconstruct the input snapshots $F(\omega^{(j)})$,
thus making the encoder learn the latent distribution accurately.
Once the generator has been trained,
we can calculate the statistics of the stochastic process using the sample paths created by the generator.
Moreover, the hyperparameters $\alpha$, $\eta$, $\gamma$, and $ \lambda $ should be chosen such that the corresponding  terms are of the same order of magnitude at the beginning of training.

\begin{algorithm}[hbt]
	\caption{Approximate the stochastic process.}
	\label{Algorithm_1}
	{\bf Input:}  The initial network parameters $\phi$, $\theta$ and  $\rho$,  the training number $ epochs $, the batch size $ N $,
	discriminator iterations $ n_D $  for each update of the encoder and the generator,
	and  the loss weights $\alpha$, $\eta$, $\gamma$  and $\lambda$.
	\begin{algorithmic}
		\For{$ t= 1, 2, \cdots, epochs$}
		\For{$ s = 1, 2, \cdots, n_D$}
		\State Sample $N$ real snapshots $\{F(\omega^{(j)})\}^N_{j=1}$.
		\State Generate fake snapshots by (\ref{F}).
		\State Use \eqref{loss_D_1} to update the parameters of the discriminator $ \mathcal{D}_\rho $ by stochastic gradient descent method.
		\EndFor
		\State \textbf{end for}
		\State Sample $N$ real snapshots $\{F(\omega^{(j)})\}^N_{j=1}$.
		\State Generate fake snapshots by (\ref{F}).
		\State Use \eqref{loss_E_1} to update the parameters of the encoder  $ \mathcal{E}_\phi $ by stochastic gradient descent method.
		\State Use \eqref{loss_G_1} to update the parameters of the generator  $ \mathcal{G}_\theta $.
		\EndFor
		\State \textbf{end for}
	\end{algorithmic}
	{\bf Output:}  The network parameters $\phi$, $\theta$ and $\rho$.
\end{algorithm}

\subsection{Solving SDEs with PI-VEGAN}
In this section, we consider the stochastic partial equation described in Section 2.1.
We use two generator networks,
denoted by $ \tilde{k}_{\theta_k}(x;z) $ and $ \tilde{u}_{\theta_u}(x;z) $,
to represent $ k(x; \omega) $ and $ u(x;\omega) $, respectively.

\begin{figure}[htb]
	\centering 
	\includegraphics[width=0.9\linewidth]{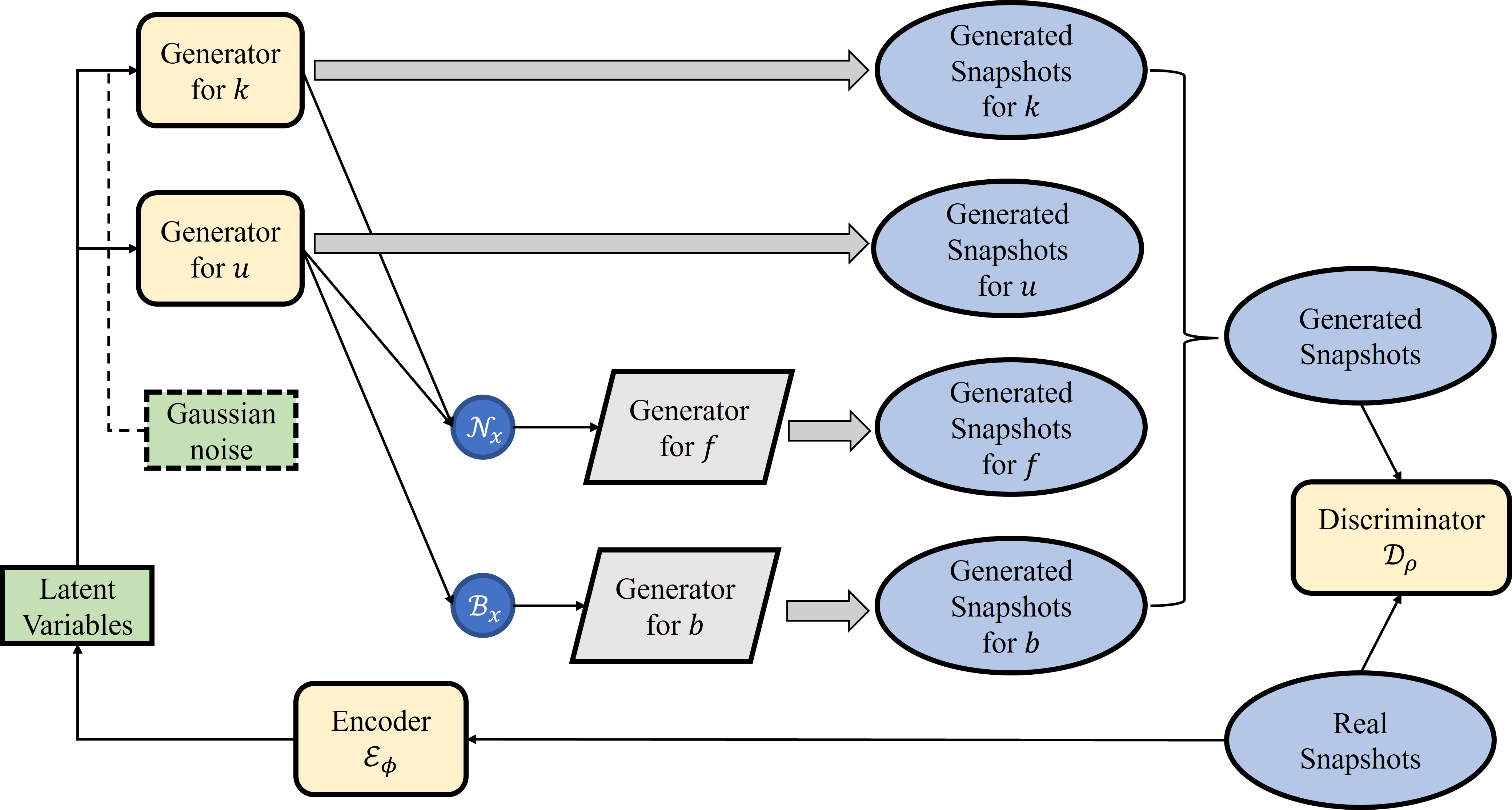}
	\caption{The architecture of PI-VEGAN for solving stochastic differential equations.
		The rounded rectangle represents the neural networks.
		The parallelogram represents the neural networks induced by differential operator $\mathcal{N}_x$ and $\mathcal{B}_x$.
		The ellipse represents snapshots from sensors, and the gray arrow represents the sampling process.
		The bracket represents concatenation.} 
	\label{PI-VEGAN} 
\end{figure}

The physical equation \eqref{eq1} is encoded into the neural network architecture by applying the differential operators $\mathcal{N}_x$ and $\mathcal{B}_x$ such that
\begin{align}
	\tilde{f}_{\theta_k, \theta_u}(x; z) &= \mathcal{N}_x [ \tilde{k}_{\theta_k}(x; z),  \tilde{u}_{\theta_u}(x; z)],
\end{align}
and
\begin{align}
	\tilde{b}_{\theta_u}(x; z) = \mathcal{B}_x [ \tilde{u}_{\theta_u}(x; z)].
\end{align}
Then we use $\tilde{f}_{\theta_k, \theta_u}(x; z)$ and $\tilde{b}_{\theta_u}(x; z) $ to model the stochastic processes
$ f(x; \omega) $ and $ b(x; \omega) $, respectively.

Now the encoder part $ \mathcal{E}_{\phi} $ takes the collected snapshots $H(\omega^{(j)})$ from \eqref{RealSnap} and returns the latent variable $ z_j $:
\begin{align}
	\label{z_sample_2}
	z_j = \mu_j + \sigma_j \odot \xi_j, \quad  \xi_j \sim N(0, I_d).
\end{align}

Then the generators $\mathcal{G}_{\theta_k} $ and $\mathcal{G}_{\theta_u} $
take the spatial coordinate $x_i$  and the learned variable $ z_j $ as the inputs,
and generate the corresponding ``fake" samples.
The generated samples are denoted by
\begin{align}
	\label{H}
	\{\tilde{H}(\omega^{(j)})\}_{j=1}^N = \{(\tilde{K}(z_j), \tilde{U}(z_j), \tilde{F}(z_j), \tilde{B}(z_j))\}_{j=1}^N,
\end{align}
where
\begin{align*}
	\tilde{K}(z_j) = (\tilde{k}_{\theta_k}(x_i^k; z_j))_{i=1}^{n_k}, \quad &\tilde{U}(z_j) = (\tilde{u}_{\theta_u}(x_i^u; z_j))_{i=1}^{n_u},\\
	\tilde{F}(z_j) = (\tilde{f}_{\theta_k, \theta_u}(x_i^f; z_j))_{i=1}^{n_f}, \quad  &\tilde{B}(z_j) = (\tilde{u}_{\theta_u}(x_i^b; z_j))_{i=1}^{n_b}.
\end{align*}
Then we feed the ``fake" snapshots and the collected real snapshots  $ \{H(\omega^{(j)})\}_{j=1}^N$  in \eqref{RealSnap} to
the discriminator $\mathcal{D}_{\rho}(\cdot)$ and perform the adversarial training on the encoder, generator and discriminator.

The final optimization objective of PI-VEGAN consists of the following three components:
\begin{align}
	\label{loss_E_2}
	\min_{\phi}\textrm{KL}(q(z|H(\omega^{(j)}))\|p(z)) + \frac{1}{N}\sum_{j=1}^{N}[\tilde{H}(\omega^{(j)})-H(\omega^{(j)})]^2,\quad p(z) \sim N(0,I_d),
\end{align}

\begin{align}
	\label{loss_G_2}
	\min_{\theta}\frac{1}{N}\sum_{j=1}^{N}\bigg(-\alpha \mathcal{D}_{\rho}(\tilde{H}(\omega^{(j)}))  + \eta [\tilde{H}(\omega^{(j)})-H(\omega^{(j)})]^2\bigg),
\end{align}

\begin{align}
	\label{loss_D_2}
	\min_{\rho}\frac{1}{N}\sum_{j=1}^{N}\bigg(\gamma [\mathcal{D}_{\rho}(\tilde{H}(\omega^{(j)})) - \mathcal{D}_{\rho}(H(\omega^{(j)}))]
	+ \lambda \mathcal{L}_{pen}(\tilde{H}(\omega^{(j)}))\bigg),
\end{align}
where the regularization term $ \mathcal{L}_{pen}(\tilde{H}(\omega^{(j)})) $ is defined similarly as in \eqref{loss_D_1}.

\begin{algorithm}[htb]
	\caption{PI-VEGAN for solving SDEs.}
	\label{Algorithm_2}
	{\bf Input:}  The initial network parameters $\phi$, $\theta=(\theta_k,\theta_u) $ and  $\rho$,  the training number $ epochs $, the batch size $ N $,
	discriminator iterations $ n_D $  for each update of the encoder and the generator,
	and the loss weights $\alpha$, $\eta$, $\gamma$  and $\lambda$.
	\begin{algorithmic}
		\For{$ t= 1, 2, \cdots, epochs$}
		\For{$ s = 1, 2, \cdots, n_D$}
		\State Sample $N$ real snapshots $\{H(\omega^{(j)})\}^N_{j=1}$.
		\State Generate fake snapshots  by (\ref{H}).
		\State Use \eqref{loss_D_2} to update the parameters of the discriminator $ \mathcal{D}_\rho $ by stochastic gradient descent method.
		\EndFor
		\State \textbf{end for}
		\State Sample $N$ real snapshots $\{H(\omega^{(j)})\}^N_{j=1}$.
		\State Generate fake snapshots  by (\ref{H}).
		\State Use \eqref{loss_E_2} to update the parameters of the encoder  $ \mathcal{E}_\phi $ by stochastic gradient descent method.
		\State Use \eqref{loss_G_2} to update the parameters of the generators $ \tilde{k}_{\theta_k} $ and $ \tilde{u}_{\theta_u} $.
		\EndFor
		\State \textbf{end for}
	\end{algorithmic}
	{\bf Output:}  The network parameters $\phi$, $\theta_u$, $ \theta_k $ and $\rho$.
\end{algorithm}

Compared with the previous PI-WGAN,
we introduce an additional optimization objective \eqref{loss_E_2} for the encoder $\mathcal{E}_\phi $ to extract the distribution information $\{z_j\} $ from the collected measurement data,
which allows the generator in our method to obtain more useful inputs and thus to learn the properties of the stochastic partial equations more accurately.
Once trained, the generator can produce approximate solutions by feeding noise from a standard normal distribution.
Algorithm \ref{Algorithm_2} outlines the implementation of PI-VEGAN,
which is suitable for forward, inverse, and mixed problems.

It is important to emphasize that random noise is not utilized in the training process but is only introduced during the testing phase.
More specifically, once the training is completed and the network parameters are determined,
random noise is fed into the generator to generate new samples that go beyond the collected training data.
By doing so, the introduction of noise does not affect the learned distribution of the collected snapshots.

\section{Numerical results}
In this section, we evaluate the effectiveness of the proposed framework numerically.
Meanwhile, we also make a comparison with PI-WGAN  under the same settings as suggested in \cite{Liu2020}.
All methods are implemented using the PyTorch framework on an Intel CPU i7-101700 platform
with 16 GB of memory and a single RTX 1070Ti GPU.

The sensors are uniformly placed in the spatial domain.
In order to obtain the training data and the reference solutions,
we simulate the sensor measurements with a Monte Carlo method using the finite difference scheme.
All the references solutions are generated from $1 \times  10^3 $ Monte Carlo sample paths.

The hyperbolic tangent function (tanh) is used as the activation function to ensure smoothness when solving higher-order derivatives.
The encoder and generator all have four hidden layers of width 128.
As the default settings, the discriminator has four hidden layers of width 128,
in addition to the four hidden layers of width 64 in Section 4.1.
The noise input to the generators follows a 4-dimensional standard Gaussian distribution.
For the hyperparameters, we use the default values $n_D = 5$, $\gamma = 1$, $\eta = 0.1$, $\alpha = 50$ and $\lambda = 0.1$,
where the loss weights are chosen such that the loss terms are of the same order of magnitude at the beginning of training.
We set the initial value of the learning rate to 0.0001, and then use cosine annealing attenuation.

Each experiment is repeated 3 times with different random seeds.
After training, we uniformly sampled 101 coordinates over the domain as validation coordinates.
We select the 30 generators at training step in the last 3000 epochs, with a stride of 100.
The average performance of the samples produced by these generators on the validation coordinates are considered.
As in \cite{Liu2020}, the following three measure are used to evaluate the performance the approach.

Wasserstein distance  \cite{Cuturi2013} is defined as  $W(p,q) = \inf_{\gamma\sim\prod(p,q)}\mathbb{E}_{x,y\sim\gamma}[\|x-y\|]$,
where $\prod(p,q)$ represents the set of all possible joint distributions combining the distributions $p$ and $q$.
The collected snapshots correspond to the real distribution, whereas the samples yielded by the generator represent the approximate distribution. 
The Wasserstein distance measures the dissimilarity between these two distributions, 
with a smaller Wasserstein distance indicating a closer match between them.

Without loss of generality, let $N$ denote the number of sample paths in a distribution, and $M$ denote the dimension of each path. 
Then the corresponding observation matrix $X$ belongs to $ \mathbb{R}^{N \times M}$.
By applying classical Principal Component Analysis (PCA) \cite{Bro2014PCA} to the observation matrix $X$, 
we can extract the principal eigenvalues that correspond to the underlying distribution.
The closer the eigenvalues between two groups of observations, the more similar the underlying distributions are.

We also evaluate the relative $L^2$ errors for the mean and standard deviation of the approximate solutions. 
The mean and standard deviation of a random function $u(x;\omega)$ can be represented by
\begin{align*}
	\mu(x) = \mathbb{E}_{\omega}[u(x; \omega)], \quad
	\sigma(x) = \sqrt{\mathbb{E}_{\omega}[(u(x; \omega)-\mu(x))^2]}.
\end{align*}
Then the corresponding relative  $L^2 $ errors of the approximation $ \hat{u}(x;\omega) $ are computed by
\begin{align}
	\frac{\|\hat{\mu}(x)-\mu(x)\|_2}{\|\mu(x)\|_2},
     \quad
	\frac{\|\hat{\sigma}(x)-\sigma(x)\|_2}{\|\sigma(x)\|_2}.
	\label{relative_error}
\end{align}

\subsection{Approximating stochastic processes}
\begin{figure}[htb]
	\centering
	\scalebox{.8}{
		\subfigure[6 sensors, $l = 0.2$]{
			\includegraphics[width=0.3\textwidth]{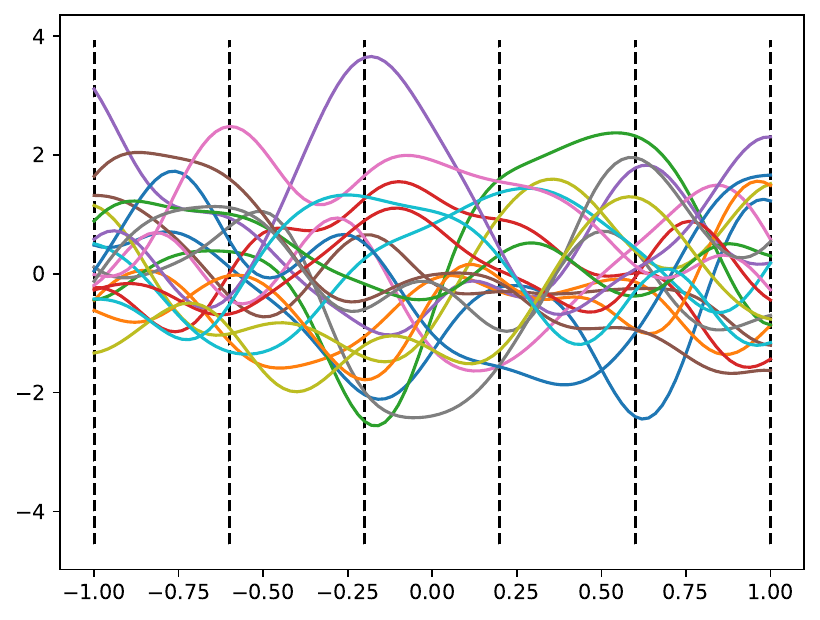}
		}
		\subfigure[6 sensors, $l = 0.5$]{
			\includegraphics[width=0.3\textwidth]{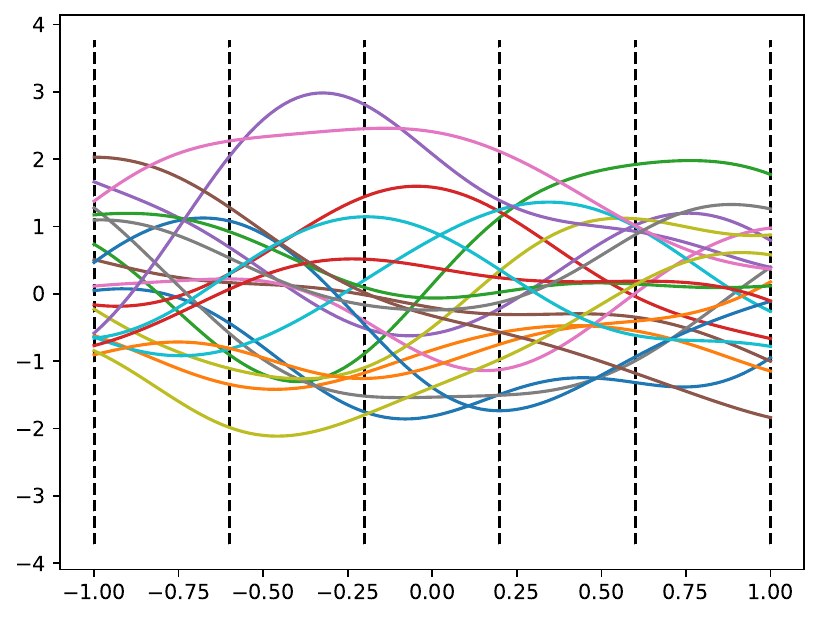}
		}
		\subfigure[6 sensors, $l = 1.0$]{
			\includegraphics[width=0.3\textwidth]{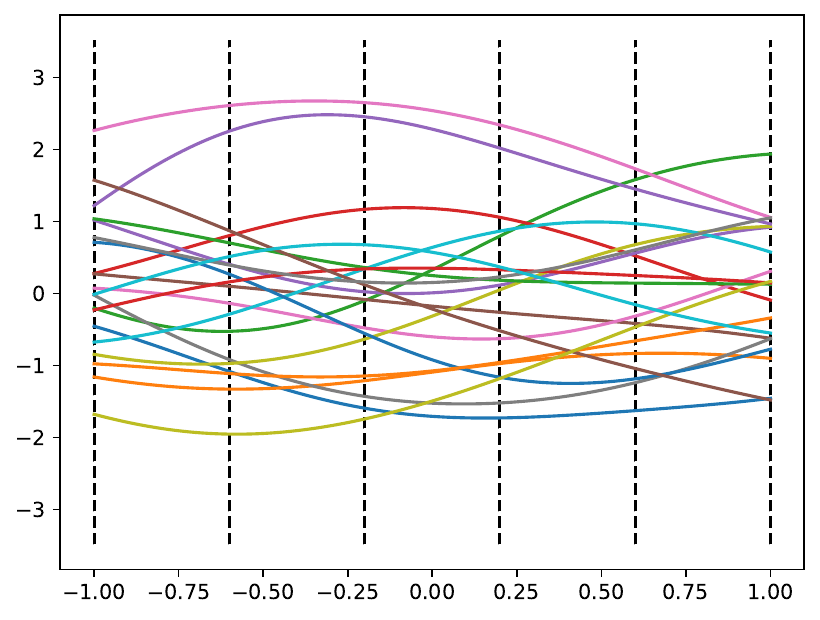}
	}}
	\\
	\scalebox{.8}{
		\subfigure[11 sensors, $l = 0.2$]{
			\includegraphics[width=0.3\textwidth]{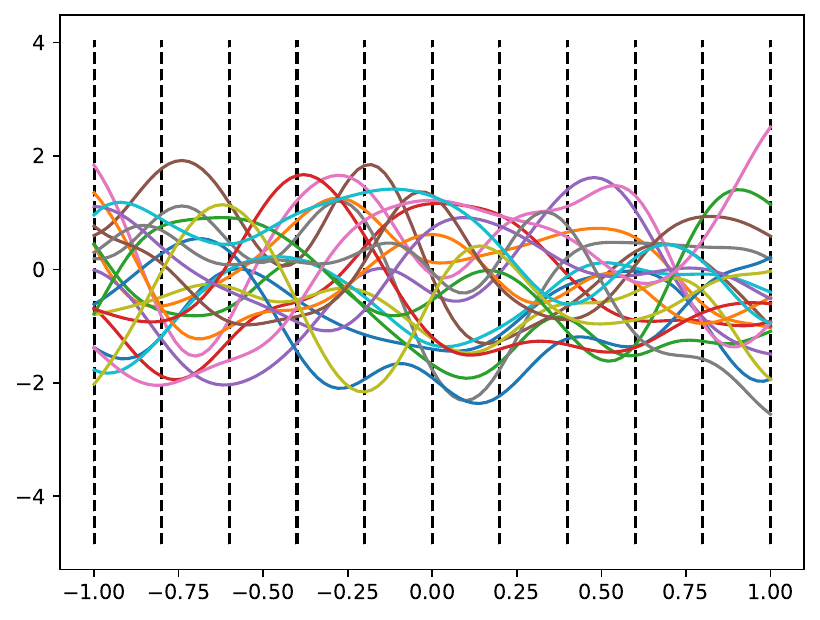}
		}
		\subfigure[11 sensors, $l = 0.5$]{
			\includegraphics[width=0.3\textwidth]{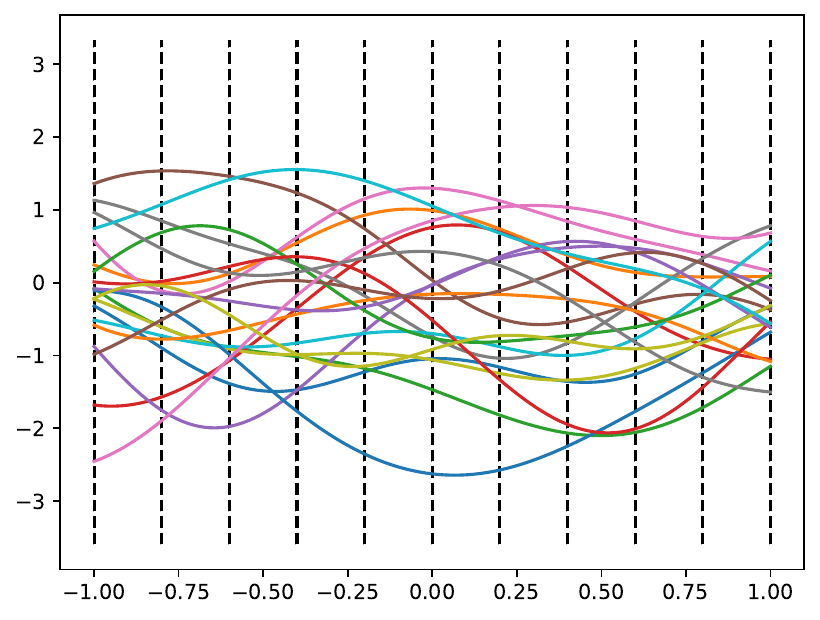}
		}
		\subfigure[11 sensors, $l = 1.0$]{
			\includegraphics[width=0.3\textwidth]{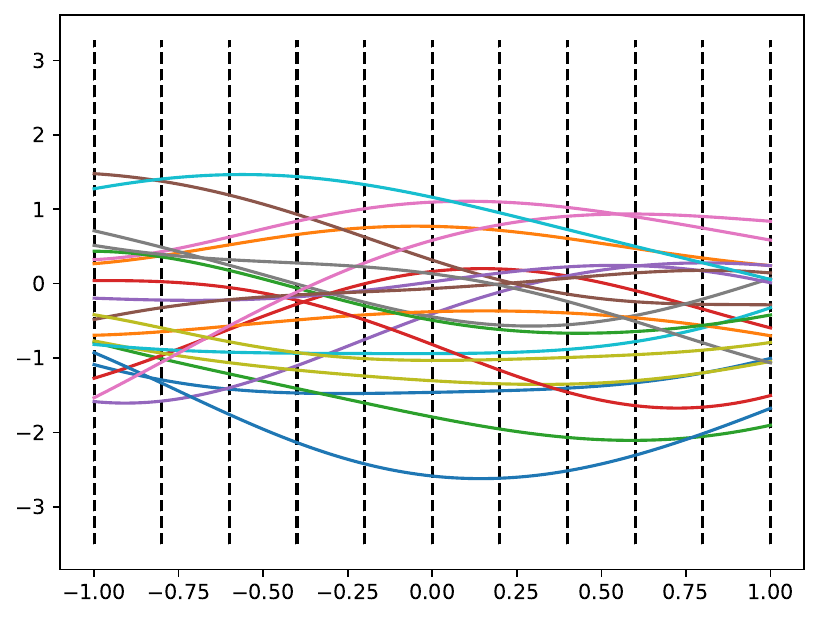}
		}
	}
	\caption{Sample paths of Gaussian processes with zero mean and squared exponential kernel.
		The positions of the sensor are indicated by black vertical dotted lines.}
	\label{sample_1}
\end{figure}

Consider the following Gaussian process with zero mean and squared exponential kernel:
\begin{align}
	f(x) \sim \mathcal{GP}(0, \exp(\dfrac{-(x-x')^2}{2l^2})),\quad x, x' \in [-1, 1],
\end{align}
where $l$ is the correlation length.  The number of sensors is set as 6 or 11.
For the correlation length, we consider $ l=0.2 $, $l= 0.5$, or  $l= 1$.
The training sample paths and the positions of the sensors are illustrated in Figure \ref{sample_1}.

\begin{figure}[htb]
	\centering
	\scalebox{.8}{
		\includegraphics[height=6cm,width=8cm]{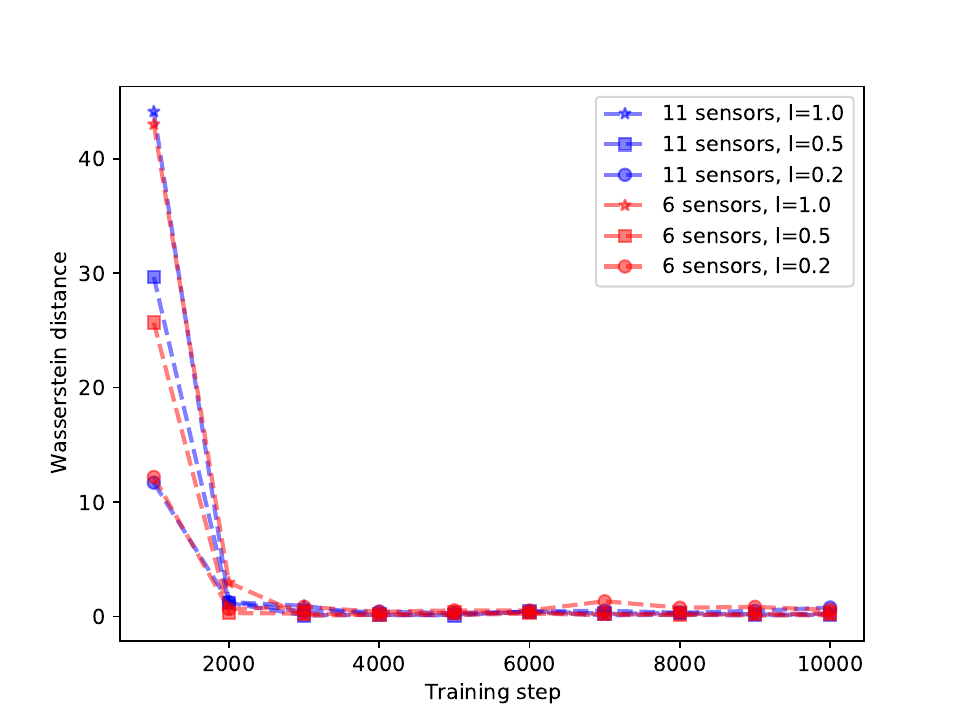}}
	\caption{Gaussian process with zero mean and squared exponential kernel: the Wasserstein distance on validation coordinates between the samples generated by our model and the reference.} 
	\label{w_distance_square_exp} 
\end{figure}

We first measure the Wasserstein distance between the samples generated by the proposed approach and the references,
It can be seen from Figure \ref{w_distance_square_exp} that the Wasserstein distance tends to zero as training proceeds,
which indicates that the distribution of the generated samples gradually approach the real distribution.
Moreover, we can find that the performance of the method are very stable for various correlation length and number of sensors.

We further compare the eigenvalues of the covariance matrix of our approach with PI-WGAN using 1000 snapshots.
In Table \ref{quantitative results_1}, we present a quantitative comparison of the first 10 eigenvalues.
The results clearly indicate that our method yields eigenvalues that are closer to the reference values in the most of cases.

\begin{table}[htb]
	\caption{
		Quantitative results of eigenvalue comparison: $l = 0.5$
	}
	\label{quantitative results_1}
	\centering
	\scalebox{0.7}{
	\begin{tabular}{|c|c|c|c|c|c|c|c|c|c|c|}
		\hline
		  &\multicolumn{10}{c|}{Eigenvalues}\\	
		\hline
		Reference
		& 49.979 & 33.914 & 12.504 & 3.979 & 1.001 & 2.04e-01 & 3.14e-02 & 4.64e-03 & 6.19e-04 & 6.60e-05\\
		\hline\hline
		PI-WGAN 6 sensors
		& 45.849 & 29.641 & 9.876 & 3.105 & 0.845 & 1.92e-01 & 9.76e-02 & 2.98e-02 & 1.59e-02 & 7.13e-03\\
		\hline
		PI-VEGAN 6 sensors
		& 45.886 & 32.796 & 11.826 & 3.382 & 0.799 & 1.54e-01 & 4.42e-02 & 1.22e-02 & 4.32e-03 & 1.61e-03\\
		\hline\hline
		PI-WGAN 11 sensors
		& 46.317 & 29.235 & 9.884 & 3.263 & 0.725 & 1.92e-01 & 7.43e-02 & 2.85e-02 & 1.20e-02 & 4.89e-03\\
		\hline
		PI-VEGAN 11 sensors
		& 46.864 & 32.263 & 11.444 & 3.570 & 0.783 & 2.08e-01 & 5.84e-02 & 2.01e-02 & 6.46e-03 & 2.19e-03\\
		\hline
	\end{tabular}}
\end{table}

\begin{figure}[!htb]
	\centering
	\subfigure[]{
		\includegraphics[height=4.2cm,width=4.9cm]{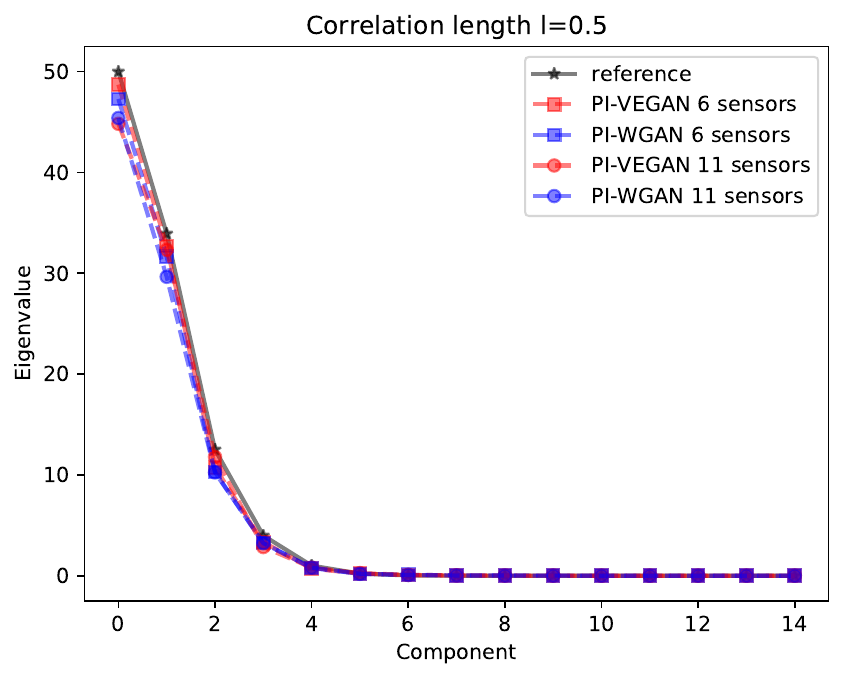}
	}
	\subfigure[]{
		\includegraphics[height=4.2cm,width=4.9cm]{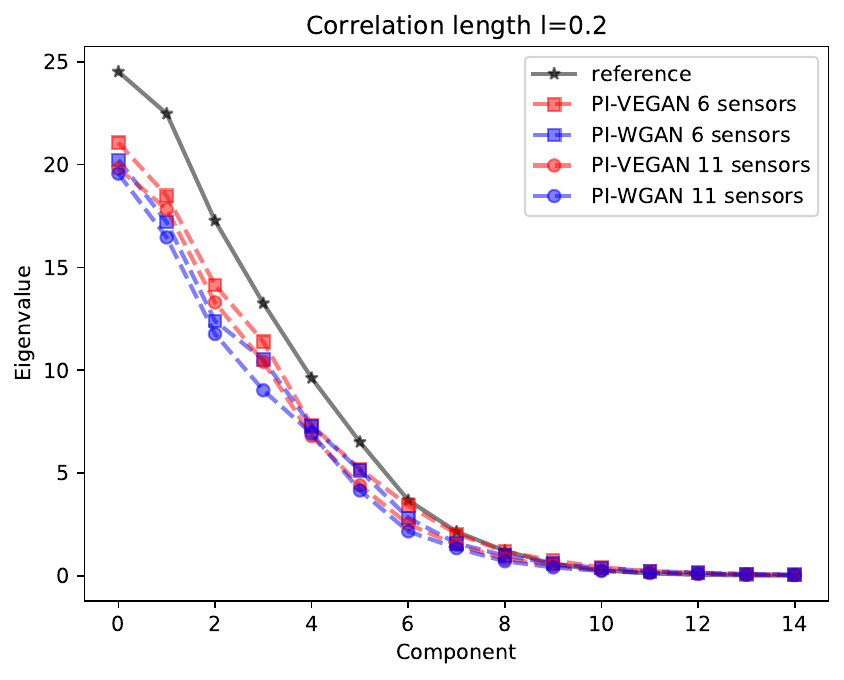}
	}
	\caption{Gaussian process with zero mean and squared exponential kernel: the eigenvalues of the covariance matrix between the samples generated by our model and the reference.
		The reference are collected by re-sampling of the stochastic processes. }
	\label{eigenvalues_square_exp}
\end{figure}

Moreover, we can observe from Figure \ref{eigenvalues_square_exp} that the eigenvalues of PI-VEGAN are closer to the reference value than those of PI-WGAN,
which means that PI-VEGAN can better capture the local behavior of high-dimensional stochastic processes.

\begin{figure}[hbt]
	\centering
	\scalebox{.8}{
		\subfigure[6 sensors, $l = 0.2$]{
			\includegraphics[width=0.3\textwidth]{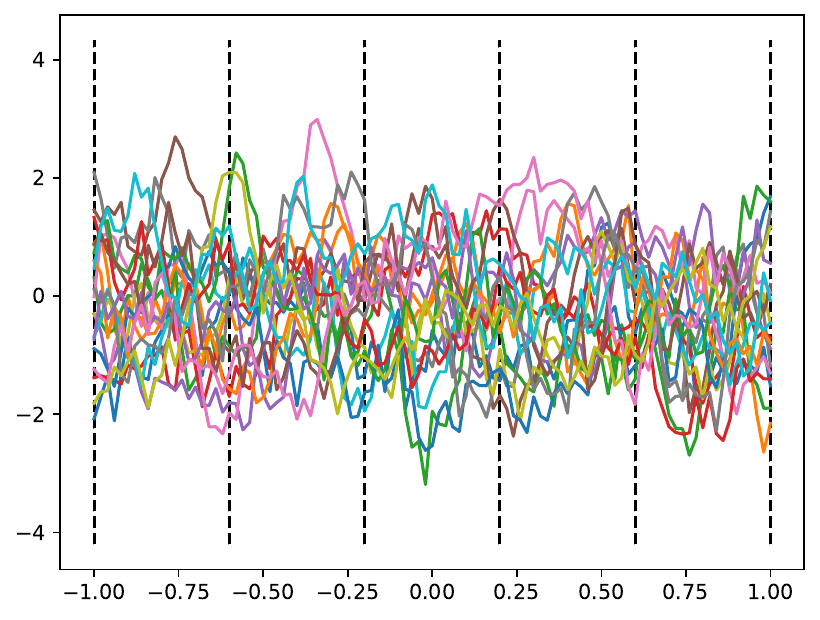}
		}
		\subfigure[6 sensors, $l = 0.5$]{
			\includegraphics[width=0.3\textwidth]{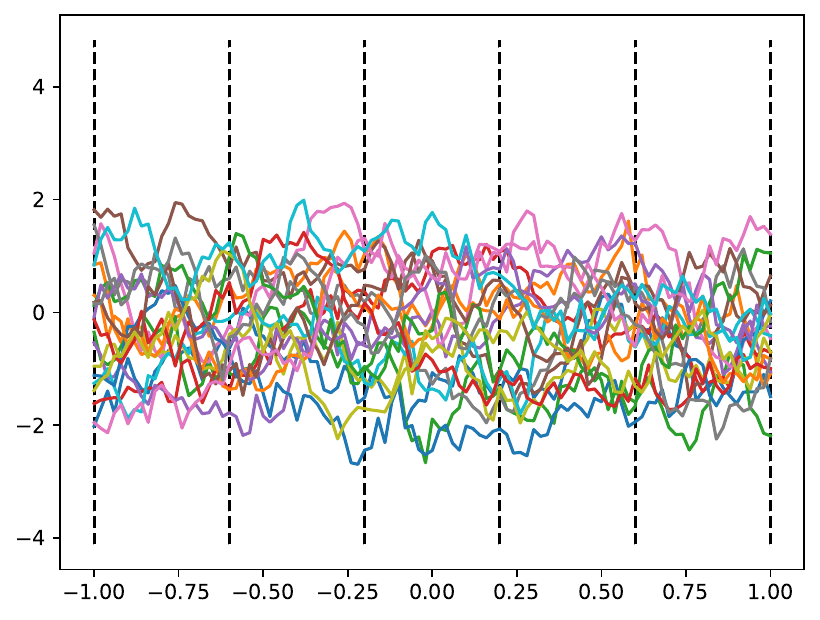}
		}
		\subfigure[6 sensors, $l = 1.0$]{
			\includegraphics[width=0.3\textwidth]{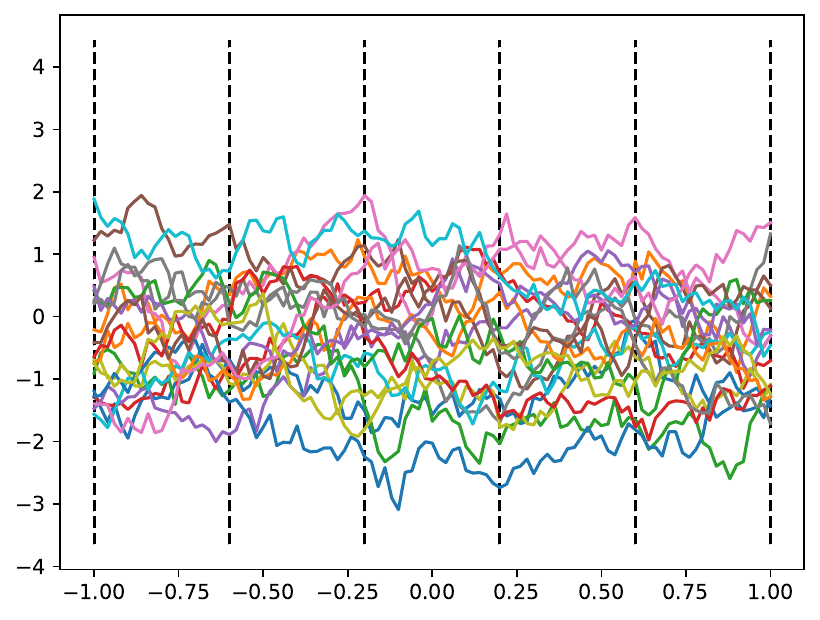}
	}}\\
	\scalebox{.8}{			
		\subfigure[11 sensors, $l = 0.2$]{
			\includegraphics[width=0.3\textwidth]{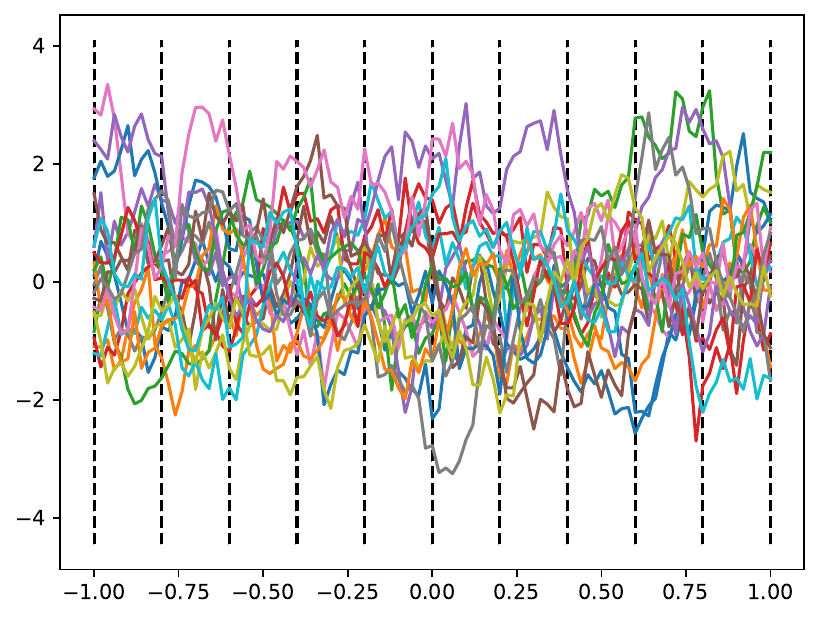}
		}
		\subfigure[11 sensors, $l = 0.5$]{
			\includegraphics[width=0.3\textwidth]{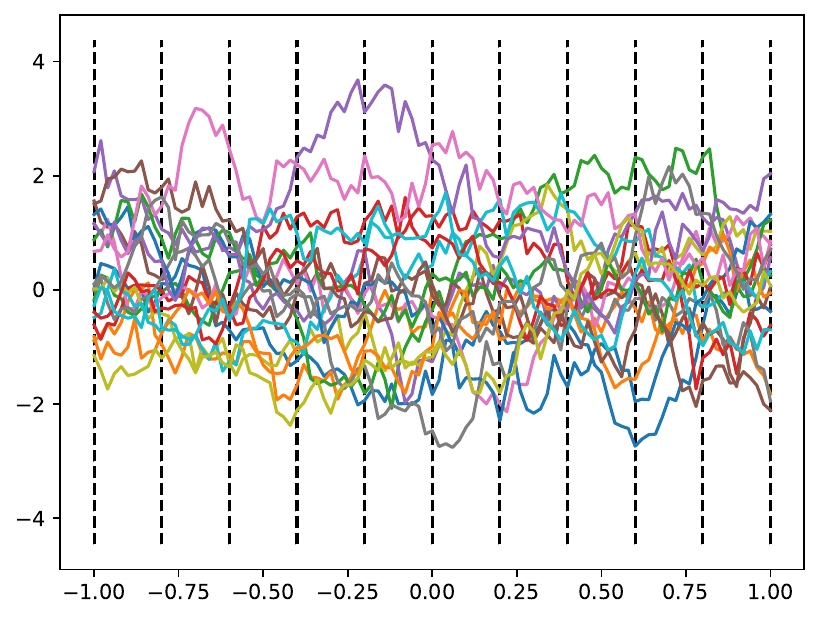}
		}
		\subfigure[11 sensors, $l = 1.0$]{
			\includegraphics[width=0.3\textwidth]{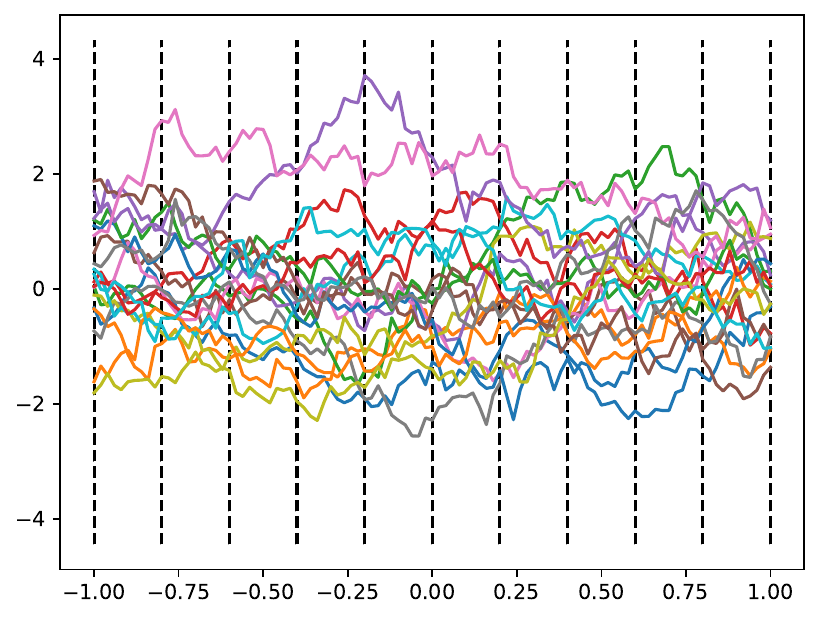}
	}}
	\caption{Sample paths of Gaussian processes with zero mean and exponential kernel. The positions of the sensor are indicated by black vertical dotted lines.}
	\label{sample_2}
\end{figure}

To further evaluate the ability in learning different statistic structures,
we consider another Gaussian process with exponential kernel:
\begin{align}
	f(x) \sim \mathcal{GP}(0, \exp(-\dfrac{|x - x'|}{l})), \quad x,x' \in [-1, 1],
\end{align}
where $l$ is the correlation length.
We test the performance of the proposed method for different number of sensors and different $l$.
The training sample paths and the positions of the sensors are illustrated in Figure \ref{sample_2}.

\begin{figure}[!ht]
	\centering 
	\scalebox{.8}{
		\includegraphics[height=6cm,width=8cm]{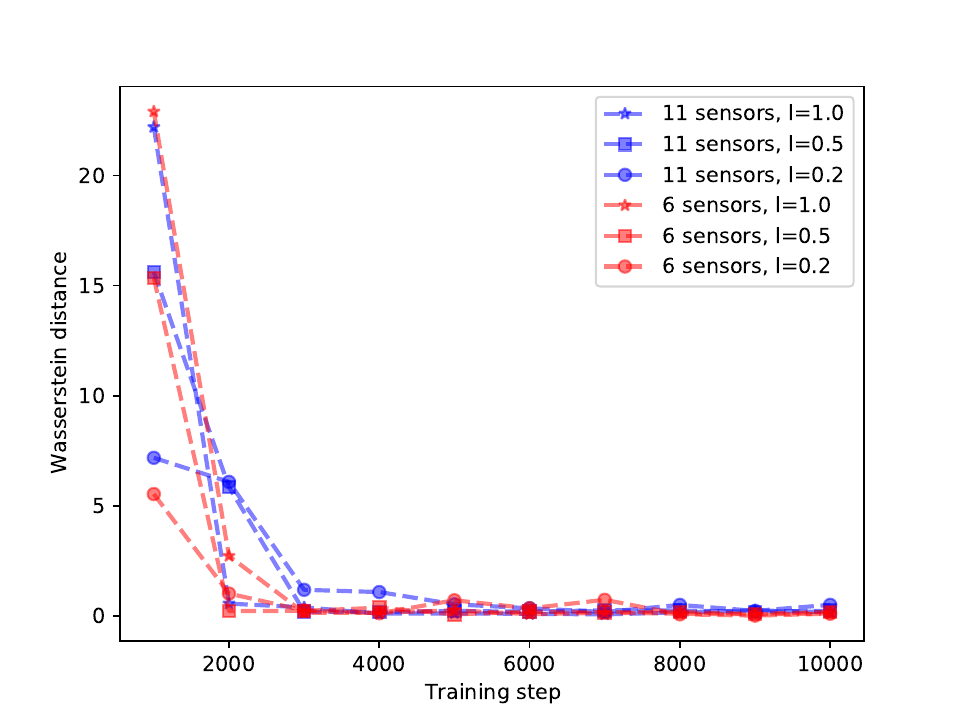}}
	\caption{Gaussian process with zero mean and exponential kernel: the Wasserstein distance on validation coordinates between the samples generated by our model and the reference.}
	\label{w_distance_exp} 
\end{figure}

\begin{figure}[!htb]
	\centering
	\subfigure[]{
		\includegraphics[height=4.2cm,width=4.9cm]{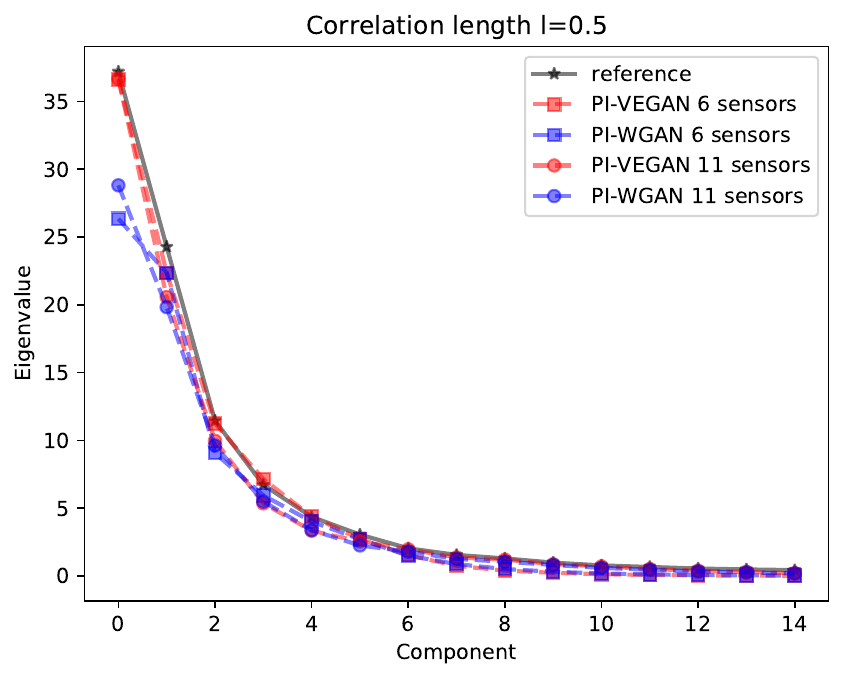}
	}
	\subfigure[]{
		\includegraphics[height=4.2cm,width=4.9cm]{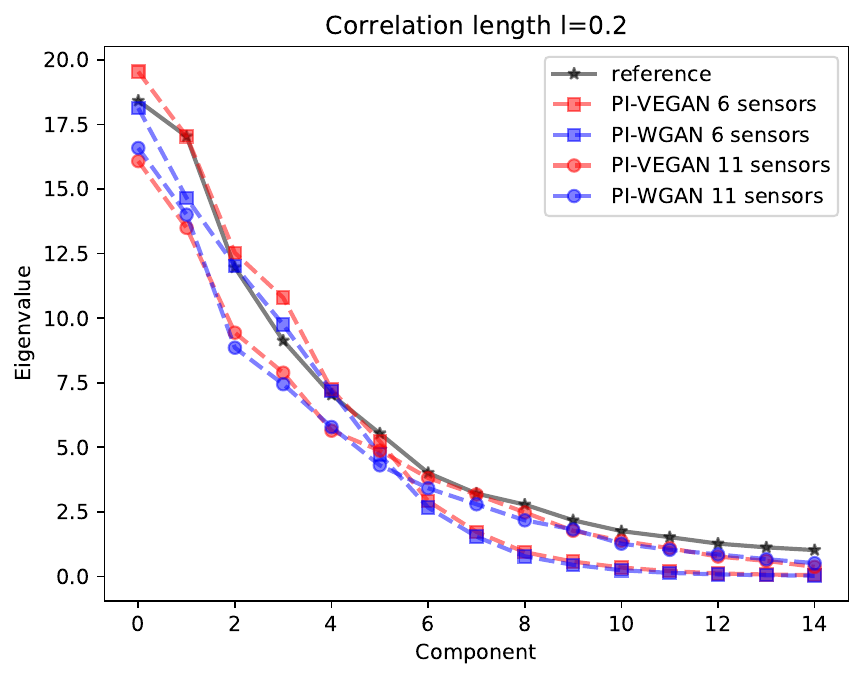}
	}
	\caption{Gaussian process with zero mean and exponential kernel: the eigenvalues of the covariance matrix between the samples generated by our model and the reference. The reference are collected by re-sampling of the stochastic processes. The above results were calculated from 1000 snapshots.}
	\label{eigenvalues_exp}
\end{figure}

As can be seen from Figures  \ref{w_distance_exp} and Figure \ref{eigenvalues_exp},
PI-VEGAN still approximates the real distribution accurately and has better performance than PI-WGAN especially in the higher-dimensional case.	

\subsection{Forward problem}
As in \cite{Liu2021}, we consider the following elliptic stochastic differential equation:
\begin{align}
	\label{SDE_example}
	-\frac{1}{10}\frac{d}{dx}[k(x; \omega)\frac{d}{dx} u(x; \omega)] = f(x; \omega), \quad x \in [-1, 1],
\end{align}
where we impose homogeneous Dirichlet boundary conditions on $ u(x; \omega$).
Let $ k(x; \omega)  $ and $f(x;\omega$) be two independent stochastic processes defined as
\begin{align}
	k(x) &= \exp [ \frac{1}{5} \sin ( \frac{3\pi}{2} (x+1) ) + \hat{k}(x) ],
\end{align}
\begin{align}
	\hat{k}(x) \sim \mathcal{GP}(0, 4/25\exp(-(x-x')^2)) ,
\end{align}
\begin{align}
	f(x) &\sim \mathcal{GP}(\frac{1}{2}, \frac{9}{400} \exp(-25(x-x')^2)).
\end{align}

\begin{figure}[htb]
	\centering
	\scalebox{.8}{
		\subfigure[sample paths of $k(x; \omega)$]{
			\includegraphics[width=0.3\textwidth]{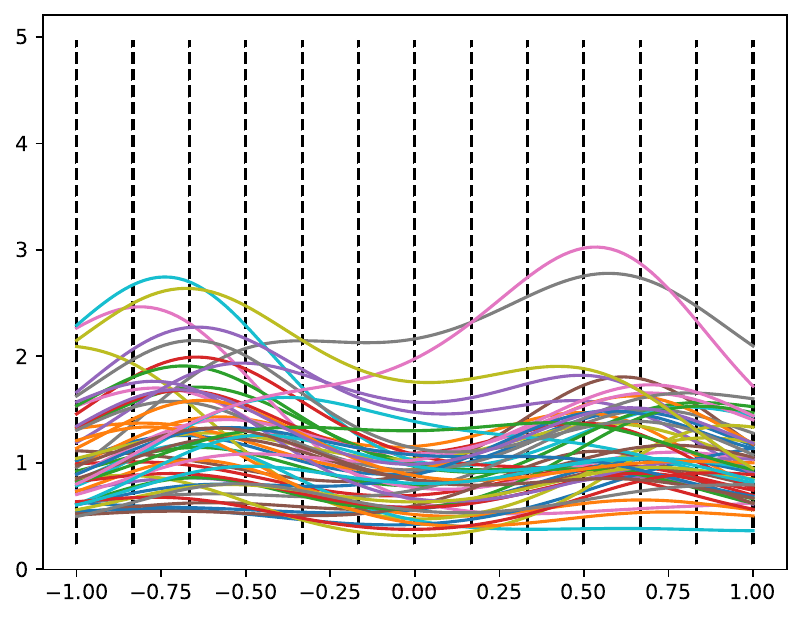}
		}
		\subfigure[sample paths of $u(x; \omega)$]{
			\includegraphics[width=0.3\textwidth]{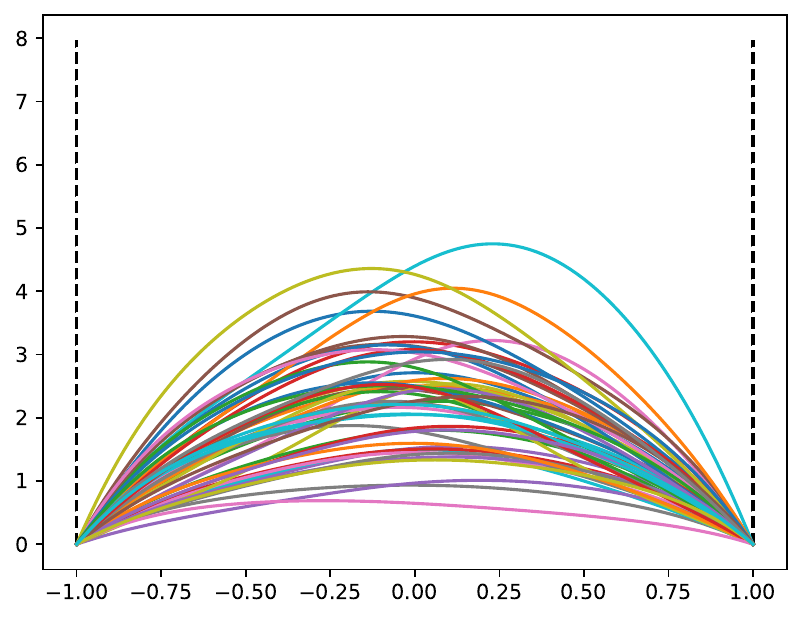}
		}
		\subfigure[sample paths of $f(x; \omega)$]{
			\includegraphics[width=0.3\textwidth]{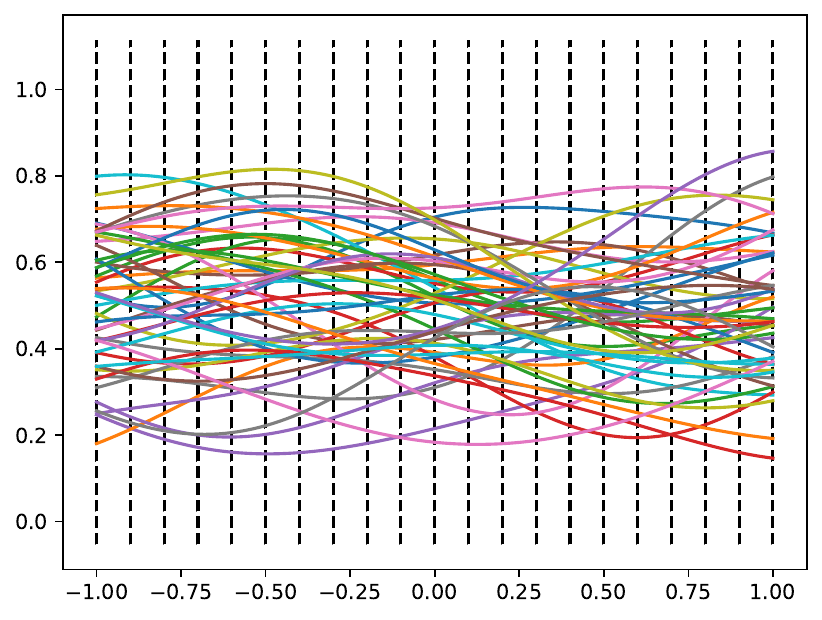}
		}
	}
	\caption{Forward problem: Sample paths of $k(x; \omega)$, $u(x; \omega)$ and $f(x; \omega)$. The positions of the sensor are indicated by black vertical dotted lines.}
	\label{sample_sde}
\end{figure}

We placed 13 sensors and 21 sensors uniformly for $k(x; \omega)$ and $ f(x; \omega)$, respectively,
and 2 sensors on the boundary to obtain the training snapshots.
The training sample paths and the positions of the sensors are illustrated in Figure \ref{sample_sde}.

\begin{figure}[htb]
	\centering
    \scalebox{.9}{
	\includegraphics[height=6cm,width=8cm]{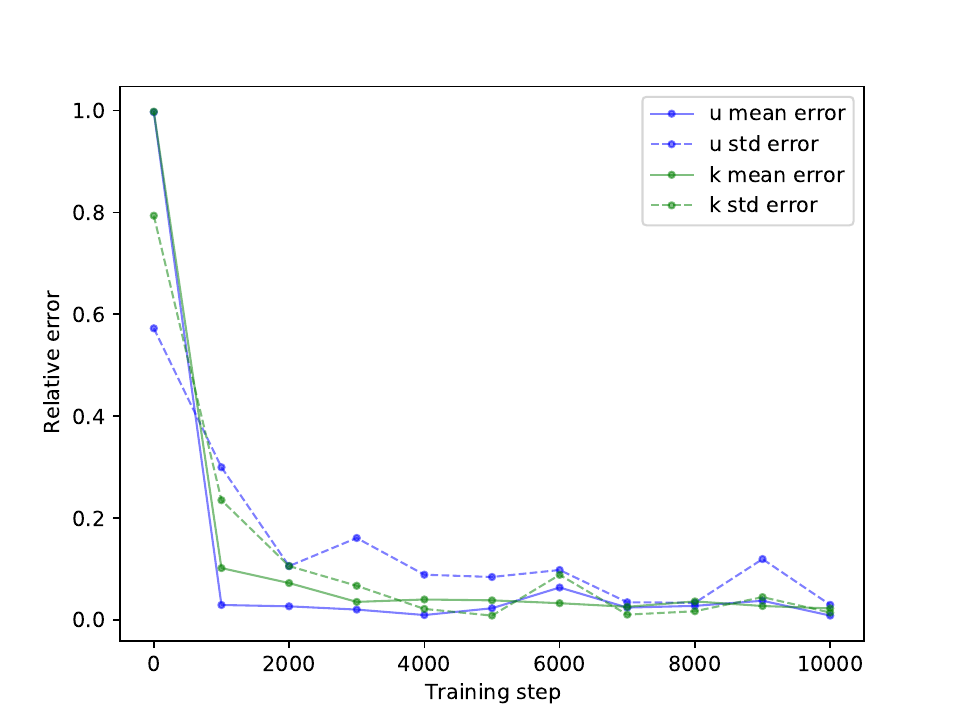}}
	\caption{Forward problem: relative error curves with various epochs.} 
	\label{forward_error_curve} 
\end{figure}

Figure \ref{forward_error_curve} shows the relative error curves of PI-VEGAN using 1000 training snapshots and a noise dimension of 4,
where the relative errors of the approximate solutions gradually approach zero as the training proceeds.		

We further investigate the influence  of the dimension of the latent variable and the number of snapshots.
To this end, we first fix the number of training snapshots to 1000 and vary the dimension of the latent variable to be 2, 4, and 20.
Then we set the latent variable dimension to 4 and vary the number of training snapshots to be 300, 1000, and 2000, respectively.
During the training process, we keep the batch size to be the total number of training snapshots.

\begin{figure}[!htb]
	\centering 
	\includegraphics[height=6cm,width=8cm]{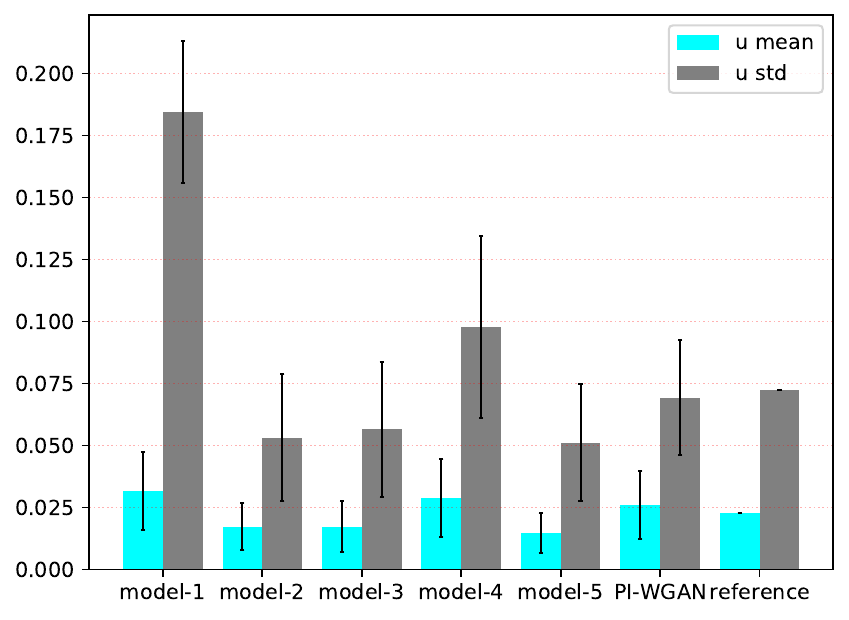}
	\caption{Relative errors (\ref{relative_error}) for the forward problem under various settings.
		(a) model-1: 1000 training snapshots and the dimensionality of the noise is 2.
		(b) model-2: 1000 training snapshots and the dimensionality of the noise is 4.
		(c) model-3: 1000 training snapshots and the dimensionality of the noise is 20.
		(d) model-4: 300 training snapshots and the dimensionality of the noise is 4.
		(e) model-5: 2000 training snapshots and the dimensionality of the noise is 4.
		(f) PI-WGAN \cite{Liu2020}: 1000 training snapshots and the dimensionality of the noise is 4.
		(g) reference: 1000 snapshots obtained by re-sampling. }
	\label{forward_error}
\end{figure}

From model-1 to model-3 in Figure \ref{forward_error},
we can find that PI-VEGAN performs better when the latent variable dimensionality becomes larger,
as  more information is encoded.
We  can further find from model-4 to model-5 that using more training snapshots leads to better performance.
Moreover, compared the model-2 with PI-WGAN,
we find that under the same conditions,
the proposed PI-VEGAN can achieve a more accurate mean value with much smaller standard deviations,
thus more stable and accurate in solving forward stochastic partial equations.

\begin{figure}[htb]
	\centering
	\scalebox{.7}{
		\subfigure[]{
			\includegraphics[width=0.5\textwidth]{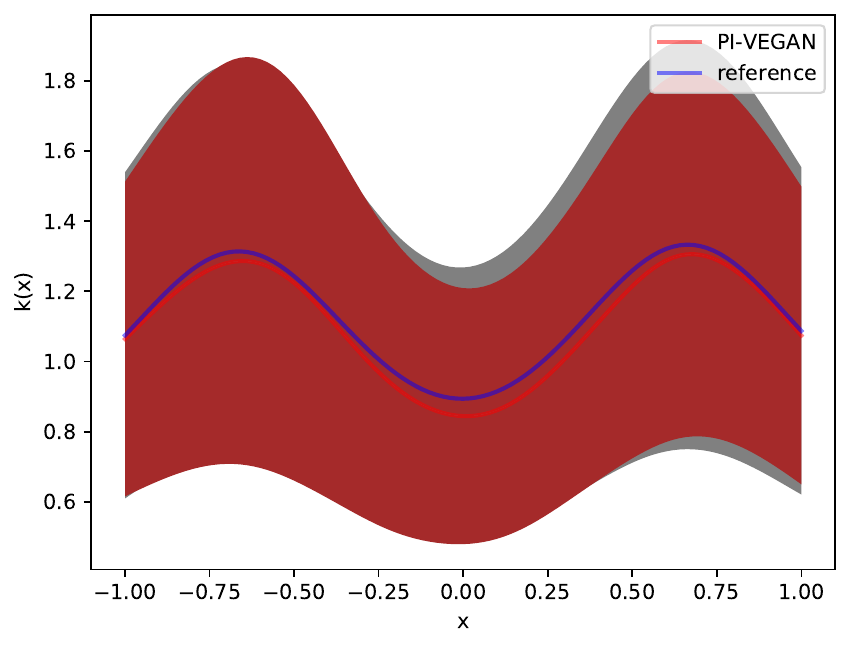}
		}
		\subfigure[]{
			\includegraphics[width=0.5\textwidth]{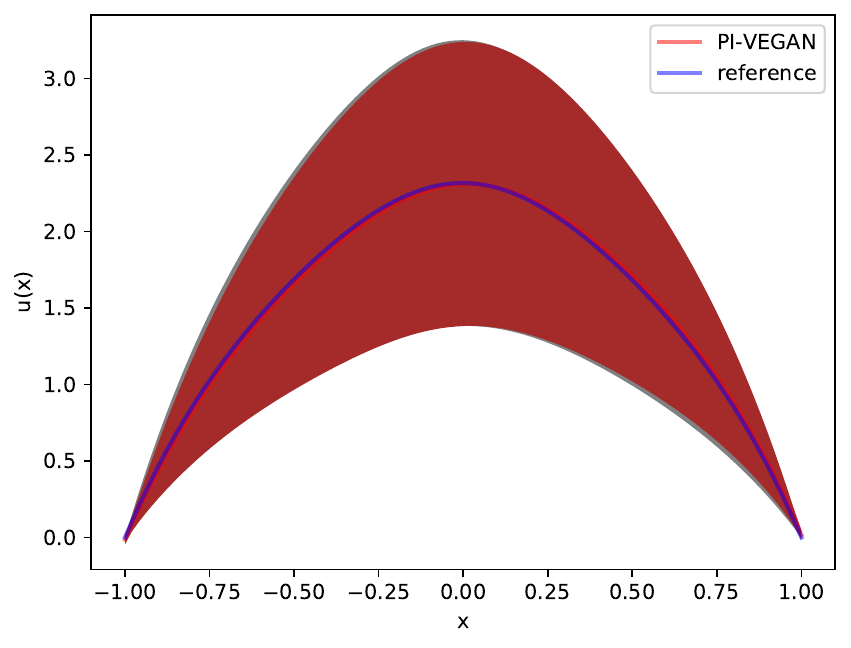}
		}
	}
	\caption{Mean and standard deviation estimate of $k(x; \omega)$ and $u(x; \omega)$ using trained model-2 are shown.
		Blue line represent the reference mean and grey shade area represents the reference standard deviation.
		Red line and brown shade area represent the mean and standard deviation of the reconstructed samples, respectively.}
	\label{distribution}
\end{figure}

Next, we use the trained model-2 to further evaluate the reproduced distribution for the process over the whole domain.
Specifically, in Figure \ref{distribution}, we show satisfactory accuracy in estimating the mean and standard deviation of the distribution.
The reference responses is calculated by obtaining another 1000 full trajectory sample paths.

\subsection{High-dimensional problem}
In this section we consider a high-dimensional stochastic equation to evaluate the ability of  PI-VEGAN in solving the problem
with the imbalance between the dimensionality of the coefficient and forcing term.
Specifically,
\begin{align}
	k(x) &= \exp [ \frac{1}{5} \sin ( \frac{3\pi}{2} (x+1) ) + \hat{k}(x) ],
	\\[5pt]
	\hat{k}(x) &\sim \mathcal{GP}(0, \frac{4}{25}\exp(-(x-x')^2)),
	\\[5pt]
	f(x) &\sim \mathcal{GP}(\frac{1}{2}, \frac{9}{400} \exp(\frac{-(x-x')^2}{a^2})),
\end{align}
where $ a $ denotes the kernel length scale of the stochastic process, with the default value $ a = 1 $.
When the correlation length of the forcing term $f(x)$ is relatively small,
the equation is referred to as a high-dimensional problem,
which allows to assess the model's capability to handle the dimensional mismatch between the forcing term $f(x)$ and the coefficient $k(x)$.

\begin{figure}[htb]
	\centering
	\scalebox{.7}{
		\subfigure[case 1]{
			\includegraphics[width=0.5\textwidth]{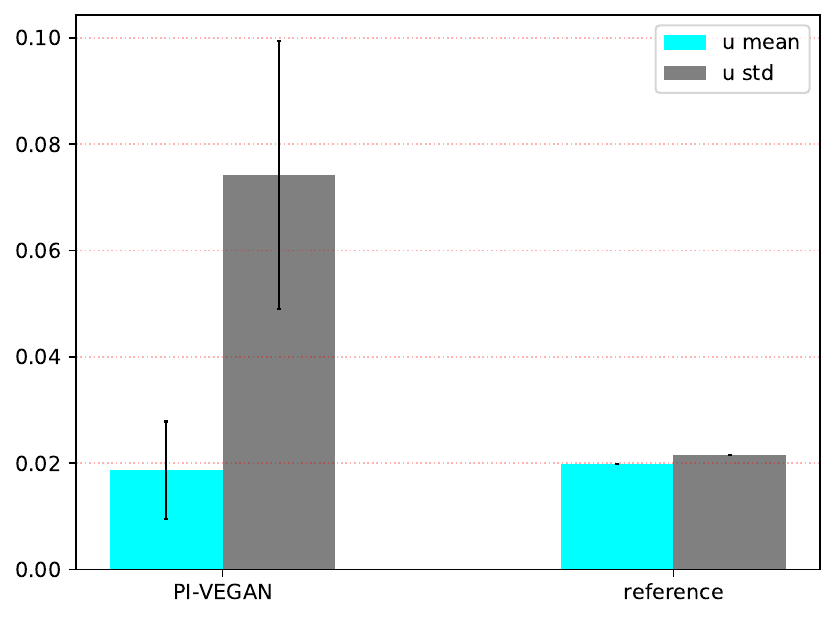}
		}
		\subfigure[case 2]{
			\includegraphics[width=0.5\textwidth]{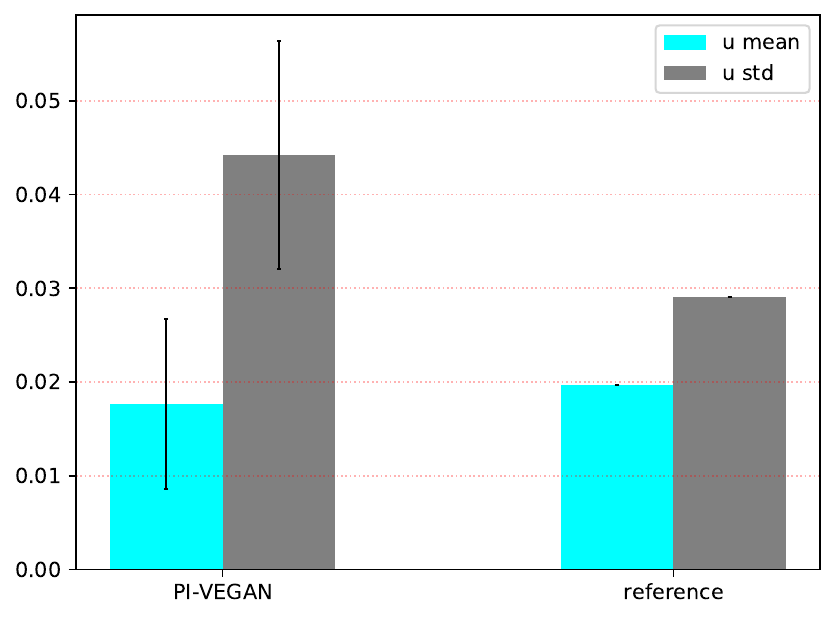}
		}
	}
	\caption{Relative errors (\ref{relative_error}) in the high-dimensional case.}
	\label{high_dimensional_error}
\end{figure}

We fixe 13 sensors uniformly for $k(x; \omega)$ and 2 sensors on the boundary of $u(x; \omega)$ to obtain 1000 training datas and consider the following two cases:
\begin{itemize}
	\item
	case 1: $a = 0.08$. We placed 21 sensors uniformly for $f(x; \omega)$ and set the dimensionality of the noise to be 10.
	\item
	case 2: $a = 0.02$. We placed 41 sensors uniformly for $f(x; \omega)$ and set the dimensionality of the noise to be 20.
\end{itemize}
The relative errors are illustrated in Figure \ref{high_dimensional_error}.

\begin{figure}[!htb]
	\centering
	\scalebox{.7}{
		\subfigure[case 1]{
			\includegraphics[width=0.5\textwidth]{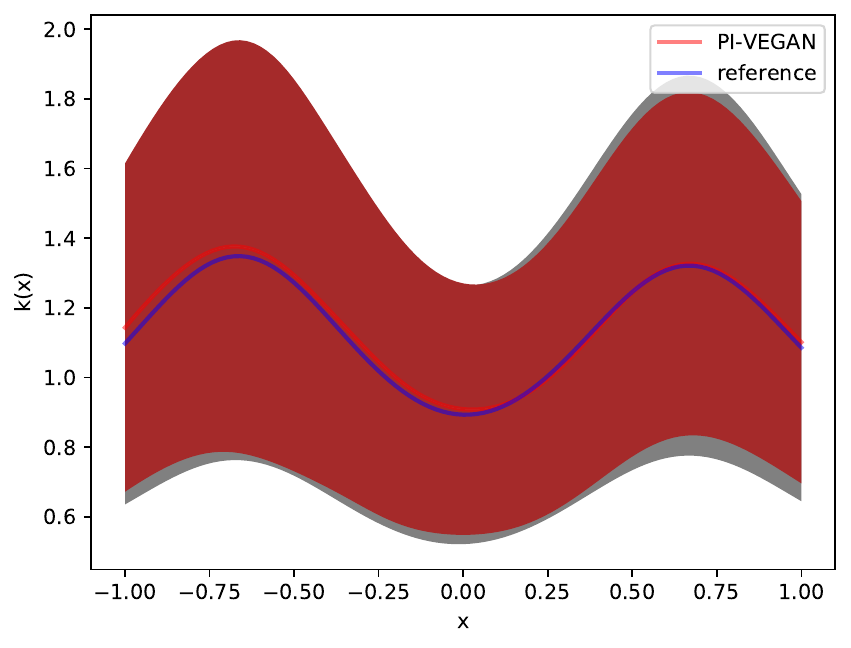}
		}
		\subfigure[case 1]{
			\includegraphics[width=0.5\textwidth]{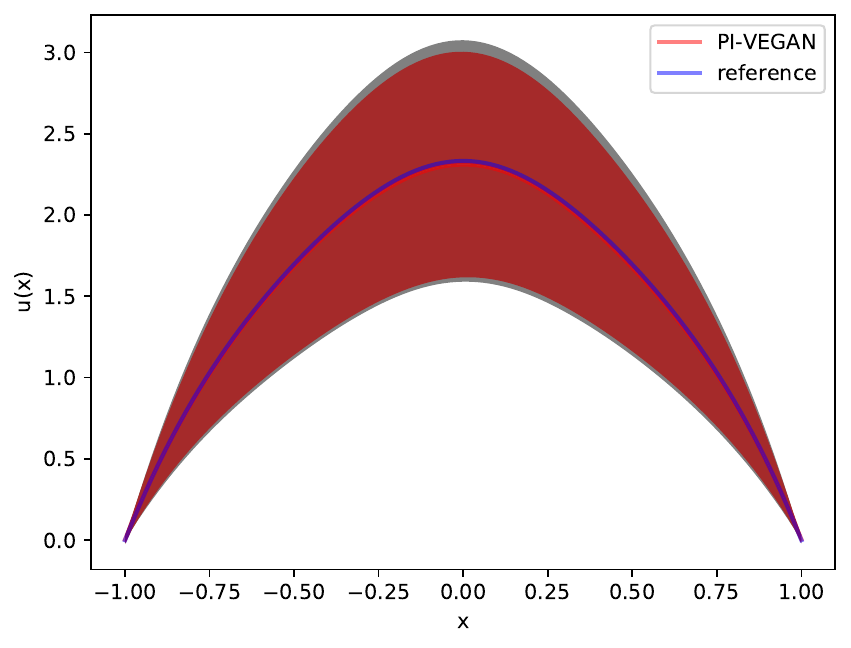}
		}}
	\\
	\scalebox{.7}{
		\subfigure[case 2]{
			\includegraphics[width=0.5\textwidth]{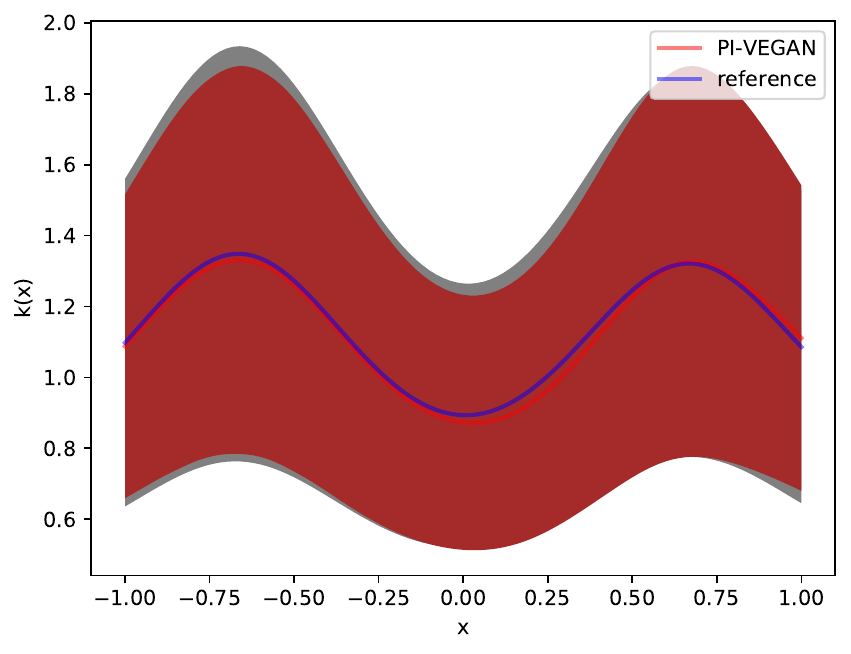}
		}
		\subfigure[case 2]{
			\includegraphics[width=0.5\textwidth]{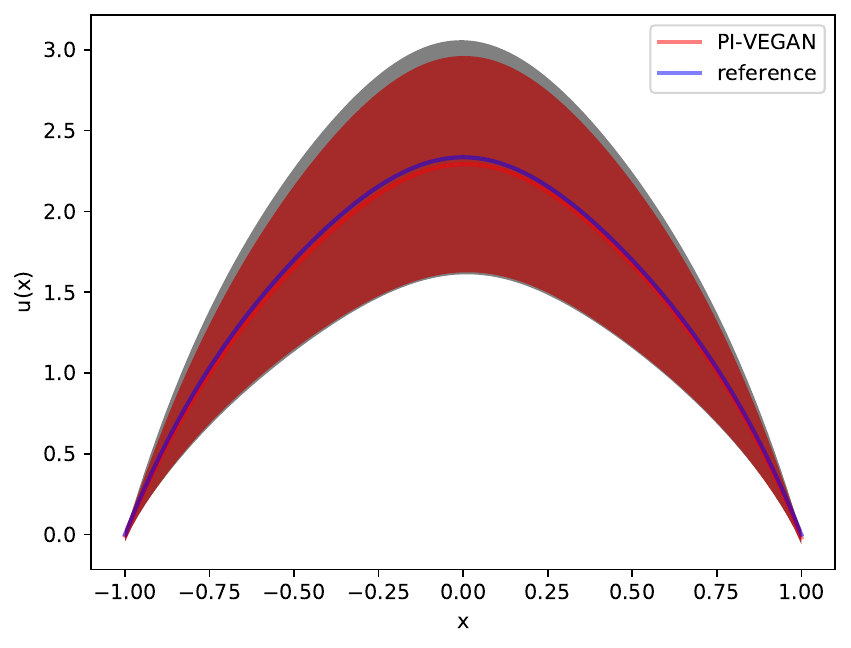}
		}
	}
	\caption{Mean and standard deviation estimate of $k(x; \omega)$ and $u(x; \omega)$ using trained model of PI-VEGAN for high-dimensional problems.}
	\label{high_dim_distribution}
\end{figure}

We can find that for the relative $ L^2 $ errors of the mean values,
the performance of PI-VEGAN is comparable to the reference method.
Although the standard deviation of the approximate solution of PI-VEGAN is higher than that of the reference method,
it is worth noting that the reference method used 101 sensors, whereas PI-VEGAN only requires a significantly smaller number of sensors.

Next, we use the trained model of PI-VEGAN to further evaluate the reproduced distribution for the process over the whole domain.
Specifically, in Figure \ref{high_dim_distribution}, we show satisfactory accuracy in estimating the mean and standard deviation of the distribution.
The reference responses is calculated by obtaining another 1000 full trajectory sample paths.

\subsection{Inverse problem}
\begin{figure}[htb]
	\centering 
	\includegraphics[height=6cm,width=8cm]{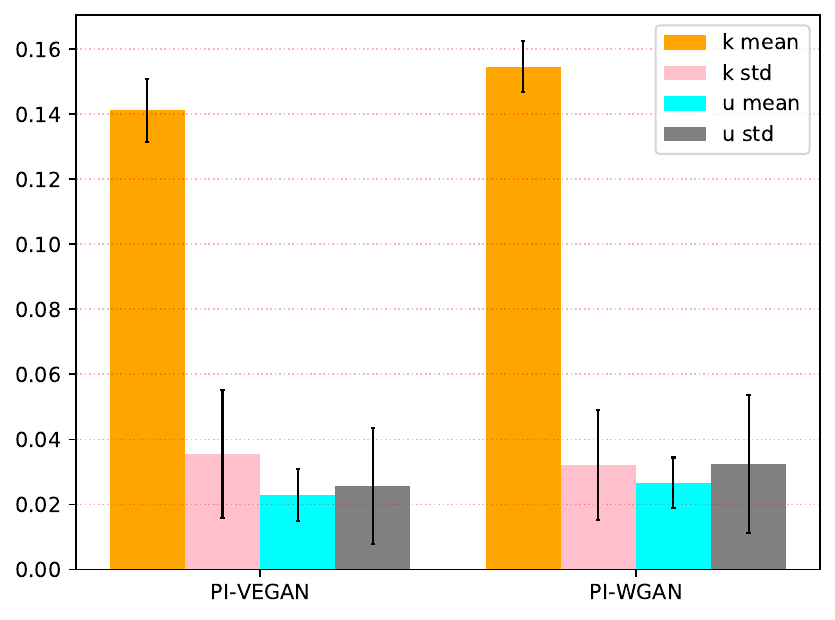}
	\caption{Relative errors (\ref{relative_error}) of PI-VEGAN and PI-WGAN for inverse problems.} 
	\label{inverse_error} 
\end{figure}

In this section we evaluate the proposed  PI-VEGAN in solving the inverse problem of the stochastic partial differential equation \eqref{SDE_example}.
To this end, we use 1 sensors, 13 sensors (including 2 on the boundary) and 21 sensors for the stochastic processes $k(x; \omega)$,
$u(x; \omega)$ and $f(x; \omega)$, respectively.
The relative errors of PI-VEGAN using 1000 training snapshots and a noise dimension of 4 are plotted in Figure \ref{inverse_error}.

In Figure \ref{inverse_error}, we show the relative error of inference from generated processes and compare it with the relative error of PI-WGAN.
Under the same conditions, the proposed PI-VEGAN can attain a more precise mean value with a comparable standard deviation.

\subsection{Mixed problem}
In this section we evaluate the proposed PI-VEGAN in solving the mixed problem of the stochastic partial equation \eqref{SDE_example}.
We fixe 21 sensors for $f(x; \omega)$ and consider the following two cases:
\begin{itemize}
	\item
	case 1: 15 sensors for $k(x; \omega)$, and 9 sensors for $u(x; \omega)$.
	\item
	case 2: 9 sensors for $k(x; \omega)$, and 15 sensors  for  $u(x; \omega)$.
\end{itemize}

\begin{figure}[htb]
\centering 
\includegraphics[height=6cm,width=8cm]{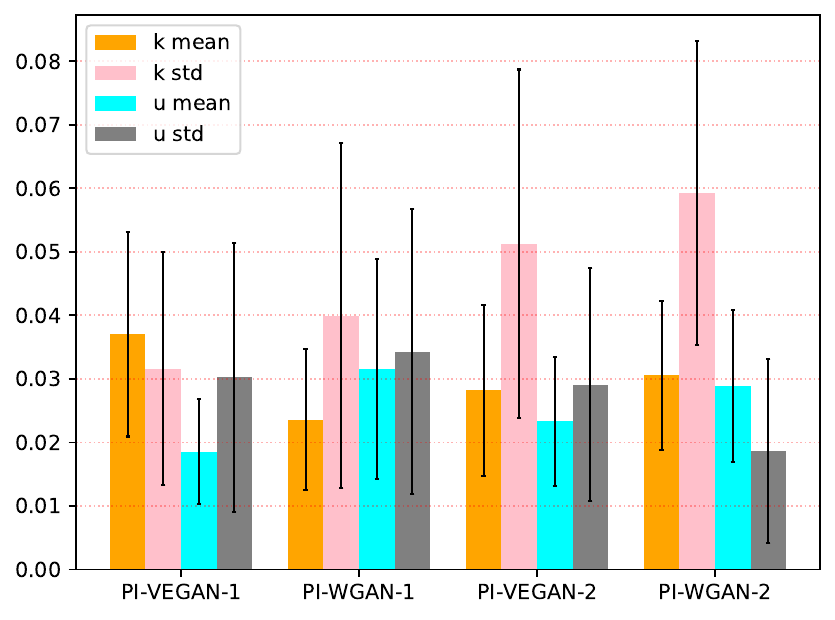}
\caption{Relative errors (\ref{relative_error}) of PI-VEGAN and PI-WGAN for mixed problems under case 1 and case 2.}
\label{mix_error} 
\end{figure}

We run PI-VEGAN and PI-WGAN under the same environments with 1000 training snapshots and a noise dimensionality of 4.
The relative errors are plotted in Figure \ref{mix_error}.
It can be found that PI-VEGAN is capable of producing smaller errors and is more stable than PI-WGAN in most scenarios.

\subsection{More comparisons}
In this section, we use the forward problem from Section 4.2 as an example to demonstrate the computational demands and training stability of the proposed method.

First, Table \ref{model_param} provides an overview of the training time, number of model parameters, and computational requirements for both PI-WGAN and our method. 
It is evident that our method lags behind in these three indicators due to the inclusion of an additional encoder.
However, it is crucial to note that this disparity is considered acceptable considering the advantages and enhancements provided by our approach.

\begin{table}[htb]
	\caption{
		The training time, number of model parameters and computation amount of PI-WGAN and PI-VEGAN.	}
	\label{model_param}
	\centering
	\scalebox{1.1}{
		\begin{tabular}{|c|c|c|c|}
			\hline
			& Time per epoch/s & Params/M & Flops/KFlops \\	
			\hline
			PI-WGAN
			& 0.405 & 0.09 & 88.19 \\
			\hline
			PI-VEGAN
			& 0.512 & 0.11 & 110.21 \\
			\hline
	\end{tabular}}
\end{table}

\begin{figure}[htb]
	\centering
	\scalebox{.8}{
		\includegraphics[height=6cm,width=8cm]{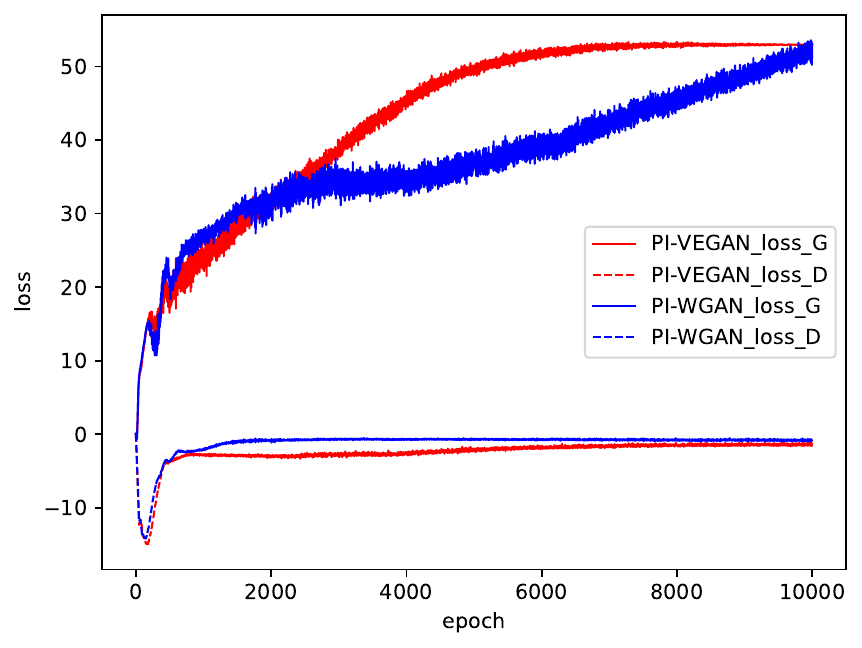}}
	\caption{Training loss curves of PI-WGAN and PI-VEGAN.} 
	\label{loss} 
\end{figure}

Next, we plot the training loss curves of PI-WGAN and PI-VEGAN in Figure \ref{loss}. 
It is evident that both our method and PI-WGAN demonstrate successful convergence of discriminator losses.
However, the generator loss curve of PI-WGAN exhibits significant fluctuations and struggles to converge.
In contrast, our generator demonstrates a more stable convergence performance.

\section{Conclusion}
In this paper, we consider the application of deep generative models in solving stochastic differential equations,
where the governing physical law is known, but information on system parameters or solution is only available at a limited number of sensors.
We proposed PI-VEGAN, a physics-informed neural network approach
that incorporates the variational encoder into generative adversarial networks, to improve the stability and accuracy of the previous PI-WGAN method.
However, the success of PI-VEGAN requires more network parameters and computational costs due to its combination of variational inference and GAN training.
Therefore, future research should focus on optimizing the model architecture for efficiency.
Furthermore, given the noisy nature of sensor data,
there remains a need for further investigation into developing generative models that account for stochastic differential equations with noisy measurements.

\section*{Acknowledgments}
This research is partially supported by National Natural Science Foundation of China (11771257) and Natural Science Foundation of Shandong Province (ZR2021MA010).

\end{document}